\documentclass{article}
\usepackage{ijcai25}

\usepackage{times}
\usepackage{soul}
\usepackage{url}
\usepackage[hidelinks]{hyperref}
\usepackage[utf8]{inputenc}
\usepackage[small]{caption}
\usepackage{graphicx}
\DeclareGraphicsExtensions{.pdf,.png,.jpg}
\usepackage{amsmath}
\usepackage{amsthm}
\usepackage{booktabs}
\usepackage{algorithm}
\usepackage{algorithmic}
\usepackage[switch]{lineno}
\usepackage{subfig}
\usepackage{bm}
\usepackage{amsfonts}
\usepackage{xcolor}
\usepackage{array}
\usepackage[T1]{fontenc}
\usepackage{tabularx} % 표의 너비를 조절하기 위해 필요함
\usepackage{booktabs} % 표를 더 깔끔하게 만들기 위한 패키지
\newcommand{\algcom}{\hfill $\triangleright$~}
\usepackage{lipsum}
\usepackage{outline}
\usepackage{multirow}

\usepackage{grffile}

\title{Automatic Curriculum Design for Zero-Shot Human-AI Coordination}

\author{
Won-Sang You$^1$
\and
Tae-Gwan Ha$^2$\and
Seo-Young Lee$^{1}$\And
Kyung-Joong Kim$^{1}$
\affiliations
$^1$Department of AI Convergence, Gwangju Institute of Science and Technology\\
$^2$School of Integrated Technology, Gwangju Institute of Science and Technology\\
\emails
\{u.wonsang0514, hataegwan, seoyoung.john\}@gm.gist.ac.kr,
kjkim@gist.ac.kr
}
\begin{document}

\maketitle

\begin{abstract}
% 제로샷 인간-AI 조정은 인간 데이터를 사용하지 않고 인간과 협력하는 ego 에이전트를 훈련시키는 것이 목적입니다. 
% 이러한 연구의 대부분은 지정된 환경에서 ego-agent의 삶과의 조정능력을 향상시키는 연구가 진행될 뿐 환경에 변화에 따른 일반화 문제를 고려하지 않습니다.

% 그러나 실제 세계에서 제로샷 인간-AI 조정이 적용될 때는 예상치 못한 환경 변화에 직면하며 환경의 특성에 따라 달라지는 사람들의 협업 능력을 고려해야합니다.
% 이미 competitive two-player games에서는 다중 에이전트 UED(Unsupervised Environment Design) 연구를 활용하여 환경 변화와 파트너 플레이어 정책을 모두 고려하여 ego-agent 학습을 진행하였습니다.
% 본 연구에서는 다중 에이전트 UED(Unsupervised Environment Design) 연구를 제로샷 인간-AI 조정 설정에 확장한 연구를 제안합니다.
% score based on return를 활용하여 환경과 co-player를 joint하게 sampling을 하고 특정 환경과 co-player pair을 과도하게 샘플링하는 것을 방지하기 위해 해밍 거리를 기반으로 환경 비유사성 score를 사용했습니다. 
% 우리는 Overcooked-AI 환경에서 인간 대리 에이전트와 실제 사람을 대상으로 zero-shot human-AI coordination performance를 평가했습니다.
% 실험에서 우리의 방법은 보이지 않는 환경에서 다른 기본 모델을 능가하고  더 높은 성능의 인간-AI 조정을 달성한다는 것을 입증했습니다.

Zero-shot human-AI coordination is the training of an ego-agent to coordinate with humans without human data. Most studies on zero-shot human-AI coordination have focused on enhancing the ego-agent's coordination ability in a given environment without considering the issue of generalization to unseen environments. Real-world applications of zero-shot human-AI coordination should consider unpredictable environmental changes and the varying coordination ability of co-players depending on the environment. Previously, the multi-agent UED (Unsupervised Environment Design) approach has investigated these challenges by jointly considering environmental changes and co-player policy in competitive two-player AI-AI scenarios. 
In this paper, our study extends a multi-agent UED approach to zero-shot human-AI coordination. 
We propose a utility function and co-player sampling for a zero-shot human-AI coordination setting that helps train the ego-agent to coordinate with humans more effectively than a previous multi-agent UED approach. 
The zero-shot human-AI coordination performance was evaluated in the Overcooked-AI environment, using human proxy agents and real humans. Our method outperforms other baseline models and achieves high performance in human-AI coordination tasks in unseen environments. The source code is available at \url{https://github.com/Uwonsang/ACD_Human-AI}
\end{abstract}

\section{Introduction}

% 강화학습의 성공적 사례, 멀티에이전트, Human-ai coodination으로 확장
% background %
Deep reinforcement learning has achieved remarkable success in diverse domains such as gaming \cite{silver2017mastering,berner2019dota,vinyals2019alphastar}, autonomous vehicles \cite{kiran2021deep}, and robotic controls \cite{fang2019curriculum}. Research in deep reinforcement learning has expanded to multi-agent reinforcement learning, where multiple agents cooperate or compete to achieve specific objectives \cite{zhang2021multi,oroojlooy2023review}. Most studies in this field have focused on developing powerful AI beyond human performance through AI-AI interactions. Recently, interest has shifted to problem-solving through human-AI interaction. 

% [Zero-shot human coordination에 대한 설명]
Some human-AI interaction studies focus on training agents that coordinate with humans in common tasks and problem-solving. These studies typically require a large amount of human data for training, but acquiring sufficient human data is challenging in real-world applications. To overcome this limitation, a zero-shot human-AI coordination study \cite{strouse2021collaborating} suggests an approach that trains the ego-agent to coordinate with humans without human data. This approach mainly uses population-based training (PBT) \cite{strouse2021collaborating,lupu2021trajectory,zhao2023maximum,lou2023pecan}.  During training, the ego-agent coordinates with various co-player agents from the population pool, rather than with a specific co-player agent. This prevents overfitting to a specific coordination strategy and helps generalize across diverse human coordination strategies.

% [UED에 대한 설명]
Unsupervised environment design (UED) \cite{dennis2020emergent} is a method that selectively curates environments for the agent by using a curriculum based on a given utility function. It curates increasingly difficult environments based on the current agent's policy. This approach allows the agent to develop the ability to generalize across all possible environments by experiencing diverse and challenging scenarios. These methods \cite{jiang2021replay,jiang2021prioritized,parker2022evolving} mainly use regret as a utility function. Recently, UED methods have been extended to competitive multi-agent settings \cite{samvelyan2023maestro}. In a multi-agent setting, the co-player policy varies depending on the environment; therefore, the multi-agent UED curates joint environment/co-player pairs for each agent. This approach has successfully trained robust agents across diverse environments and co-players. By extending this method to the zero-shot human-AI coordination setting, we could train a robust agent that can coordinate with real humans across diverse environments.

However, existing approaches in zero-shot human-AI coordination only focus on unseen co-players and do not consider generalization to unseen environments. Real-world applications of human-AI coordination face various unpredictable variables and situations, including diverse partners and changing environmental conditions. Therefore, it is essential to train the ego-agent to be robust against both diverse co-players and environments. While UED methods could potentially address this challenge by extending to zero-shot human-AI coordination settings, the existing multi-agent UED method is not directly applicable to coordination tasks. This method was developed for competitive zero-sum games and does not match the collaborative nature of human-AI coordination tasks. This mismatch motivates the need for a specialized approach that combines the benefits of zero-shot coordination and unsupervised environment design while addressing their respective limitations.

\begin{figure*}[ht]
    \centering
     \includegraphics[width=0.9\textwidth]{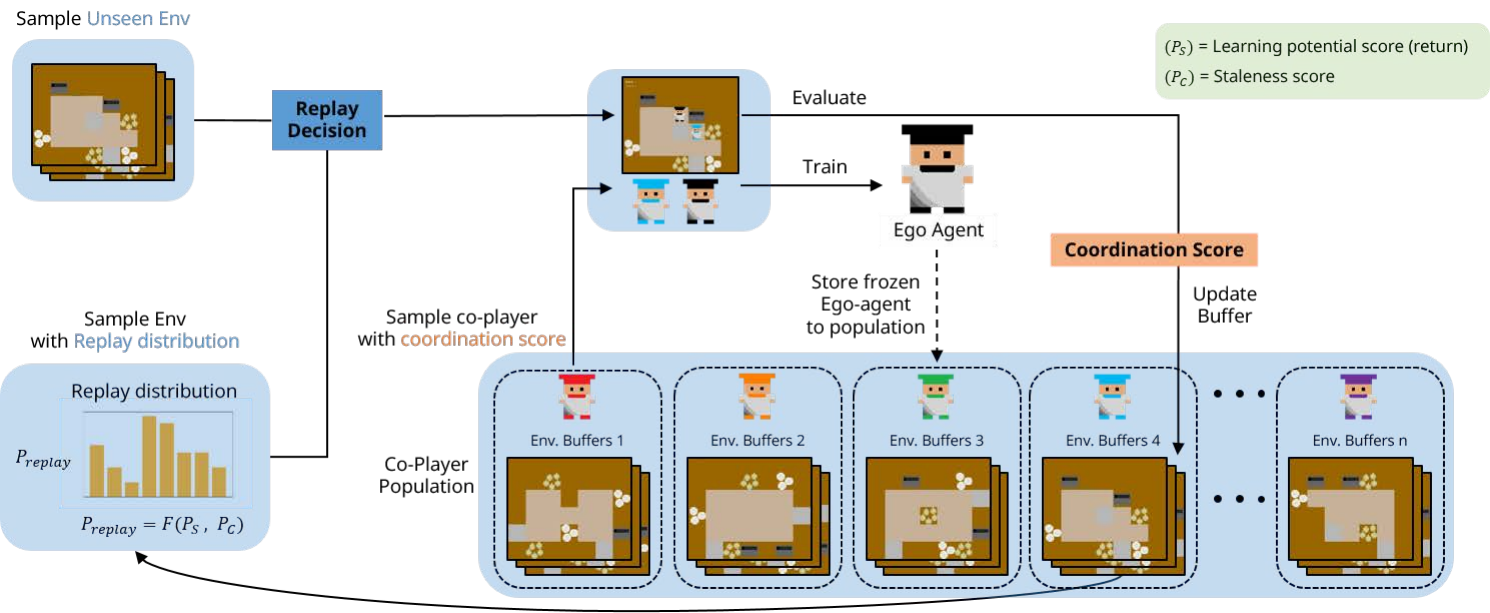}
    \caption{\textbf{Overview of the proposed method.} The ego-agent is trained using a population of co-players whose individual environment buffers maintain a fixed size of \textit{K}, containing the environments with the lowest coordination scores. Our approach samples the co-player based on their coordination scores within the population. Depending on the replay decision, the environment is either sampled from the unseen environment buffer or the co-player’s buffer, guided by a replay distribution that prioritizes environment sampling based on coordination scores. After sampling, our method calculates the return from the episode trajectory within the sampled environment, which is used to compute the coordination score. This coordination score is then updated within the co-player's buffer.}
    \label{fig:overall}
\end{figure*}

To address these problems, we introduce the automatic curriculum design for zero-shot human-AI coordination. We adopt a return-based utility function and co-player sampling for zero-shot human-AI coordination. The return measures the ego-agent's coordination performance with co-players. It helps the agent learn better coordination strategies even in challenging scenarios and with difficult co-players. The contributions of our study are as follows: (1) We propose a multi-agent UED framework for zero-shot human-AI coordination that uses a return-based utility function and co-player sampling strategy. (2) We show that our method demonstrates higher coordination performance with human proxy agents and real human partners compared to other baselines. Figure \ref{fig:overall} shows an overview of our proposed method. 

The rest of the paper is organized as follows. Section \ref{sec:Related Works} provides a review of related work in unsupervised environment design and zero-shot human-AI coordination. Section \ref{sec:Method} presents our proposed automatic curriculum design method, including the return-based utility function and the co-player sampling strategy. Section \ref{sec:Setup} describes the experimental setup and evaluation metrics. Section \ref{sec:Result} presents the experimental results with human proxy agents and real human participants. Finally, Section \ref{sec:Conclusion} concludes the paper and discusses future research directions.

\section{Related Works}
\label{sec:Related Works}
\subsection{Unsupervised Environment Design Strategies}
% 커리큘럼, UED를 합친 버젼의 Related work입니다

Unsupervised Environment Design (UED) is a method in which the teacher curates a curriculum to enable the student agent to generalize across all possible levels of the environment based on a utility function. The simplest version of the UED teacher uses the utility function as a constant $C$ for uniform random sampling of the environment level $U_t(\pi, \theta) = C$. Here, $t$ is the teacher, $\pi$ is the student's policy, $\theta$ is the environment level parameter \cite{jakobi1997evolutionary}. Most UED studies use regret as a teacher's utility function, defined as the difference between the expected return of the current policy and the optimal student policy. The teacher aims to maximize this regret \cite{dennis2020emergent,gur2021environment}, which can be defined as : $U_t(\pi, \theta)  =\max_{\pi^* \in \Pi}\{\textsc{Regret}^{\theta}(\pi,\pi^*)\} = \max_{\pi^* \in \Pi}\{V_\theta(\pi^*)-V_\theta(\pi)\}$, where $\pi^*$ is the optimal policy in $\theta$, $\Pi$ is the set of student policies. This regret-based utility function encourages the teacher to curate challenging environments for the student agent. Through the training process, the student's policy will converge to the minimax regret policy by minimizing regret in teacher-curated environments \cite{dennis2020emergent}: $\pi \in \arg\min_{\pi \in \Pi}\{\max_{\theta,\pi^* \in \Theta, \Pi}\{\textsc{Regret}^{\theta}(\pi,\pi^*)\}\}$ where $\Theta$ is the set of environment- level parameters. Since the optimal policy $\pi^*$ cannot be accessed in practice, the regret must be approximated.

Several UED studies have explored the use of regret. Prioritized Level Replay (PLR) \cite{jiang2021prioritized} and Robust PLR \cite{jiang2021replay} leverage a regret-based utility function to train the agent by selectively sampling environments with high learning potential.
ACCEL \cite{parker2022evolving} continuously edits the environment that pushes the boundaries of the student agent's capabilities by combining an evolutionary approach with Robust PLR.
ADD \cite{chungadversarial} generates diverse and challenging environments using diffusion models with entropy regularization, while No Regrets \cite{rutherford2025no}  addresses limitations of regret approximation by introducing Sampling for Learnability, which prioritizes environments where agents succeed approximately 50$\%$.

MAESTRO \cite{samvelyan2023maestro} extends single-agent UED to multi-agent UED in a competitive setting by curating environment/co-player pairs while considering the environment-dependent co-player policy. Recently, the Overcooked Generalisation Challenge (OGC) \cite{ruhdorfer2024overcooked} highlighted generalization challenges in human-AI coordination, showing that existing UED-based methods struggle to achieve zero-shot coordination with novel partners and environments.

However, most existing methods are primarily effective in single-agent and competitive settings. Regret-based utility functions are not suitable for human-AI coordination tasks, where coordination and common rewards are essential, as they do not directly reflect collaborative success with diverse partners. In this paper, our method extends competitive multi-agent UED to a zero-shot human-AI coordination setting by using an appropriate utility function.

% UED는 강화 학습 환경의 구조를 직접 수정하는 비교적 최근의 접근 방식입니다. UED는 앞선 연구들처럼 단순히 적합한 환경을 선택하는 것이 아니라 agent의 학습 진행 상황에 맞게 환경을 동적으로 설계하거나 조정합니다.
% UED가 제공하는 학습 환경은 에이전트가 조기에 과도한 도전에 노출되거나 학습 자극이 부족하여 정체되지 않도록 함으로써 일반화 능력을 향상시키는 것으로 나타났습니다. 따라서 agent의 일반화 능력을 향상시키기 위해 UED를 사용하는 데 많은 연구 노력이 기울여지고 있습니다.

% 서로 다른 맥락의 여러 UED 연구가 존재합니다. 
% PLR은 GAE 점수를 사용하고,  Robust PLR은 positve value loss를 사용해서 학습 잠재력이 높은 환경을 선택적으로 샘플링합니다. 
% ACCEL은 후회 기반 목표함수를 사용하는 환경에서 진화 원리에 기반한 알고리즘을 도입함으로써, 에이전트의 능력 한계에 도전하는 새로운 환경을 지속적으로 생성합니다.
% MAESTRO는 같은 환경에서도 co-player에 따라 학습 잠재력의 변화함을 확인하였고, 환경/co-player 쌍을 샘플링함으로써, 기존의 UED를 멀티 에이전트 경쟁 환경으로 확장합니다. 
% 우리의 방법은 인간-AI 협력 환경을 평가하는 더 적합한 메트릭을 사용하여, 기존 UED를 다중 에이전트 협력 환경으로 확장합니다.

\subsection{Zero-Shot Human-AI Coordination}
In multi-agent cooperation tasks, achieving generalization to partner agents is an important problem. Zero-shot coordination \cite{hu2020other} and ad hoc team cooperation \cite{stone2010ad} aim to address this problem. These approaches commonly rely on self-play methods in which an agent learns through collaboration with its clone. However, self-play methods suffer from limitations. Because they assume partner agents to be optimal or similar to themselves during training, their performance is poorly coordinated with unseen partners such as humans or other agents \cite{carroll2019utility}.

To address these issues, \cite{devlin2011empirical} increases partner diversity and encourages self-play agents to learn different strategies.
One of these approaches is Population-Based Training (PBT), which leverages diversity within a population of partner agents. In each iteration, agents are paired with different partners in the population, allowing them to learn various strategies simultaneously. Fictitious Co-Play (FCP) \cite{strouse2021collaborating}  is a well-known PBT-based approach. It uses the population of past checkpoint self-play agents as partners to train the ego-agent. Trajectory diversity (TrajeDi) \cite{lupu2021trajectory} improved performance in cooperative tasks by generating diverse policies within the population. It used the Jensen-Shannon divergence to measure the diversity between different policies. Similarly, MACOP \cite{yuan2023learning} focused on expanding partner diversity by continually generating a wide range of incompatible teammates through evolutionary processes. Furthermore, MEP \cite{zhao2023maximum} was proposed to mitigate potential issues arising from changes in the distribution of self-play agents during interactions with unseen partners. Recent approaches have further advanced zero-shot coordination. E3T \cite{yan2023efficient} introduces an efficient end-to-end training framework using a mixture partner policy to eliminate the need for pre-trained populations. CoMeDi \cite{sarkar2023diverse} leverages cross-play to encourage greater diversity than traditional statistical approaches and achieve generalized cooperation.

In addition to these diversity-driven approaches, recent work has explored the use of large language models (LLMs) for human-AI coordination. HAPLAN \cite{guan2023efficient} improves coordination by establishing preparatory conventions using LLM before interaction. HLA \cite{liu2024llm} employs a hierarchical framework that combines high-level reasoning with low-level control for real-time coordination. 

However, most of these approaches focus on improving coordination with unseen partners through partner diversity and do not explicitly address the challenges posed by unseen environments. In this study, we extend the focus to coordination across diverse unseen environments by constructing a population of co-player agents trained in various settings. Through this approach, we seek to address the poor coordination between AI agents and human partners in unseen environments.

%이러한 문제점을 해결하기 위해 SP에 대해 알고리즘 적 변화를 주어 agent가 더 높은 보상으로 수렴하게 한 시도가 있다[ref]. 최근의 접근법은 Population-based training로 여러 개의 에이전트로 구성된 Population을 제작하고, 해당 Population에 포함된 agent들을 병렬적으로 학습 시키는 방법이다[ref]. 해당 Population은 서로 다른 agent로 구성 되어있기 때문에 각 agent들이 환경에 더 잘 적응하면서 다양한 전략을 동시에 학습 할 수있 게 한다. N개의 self-play agent와 self-play agent 들의 과거 checkpoint들을 교차 학습 시키는 Fictitious Co-Play를 제안하여 협력을 목표로 하는 게임 환경에서도 Self-Play를 기반으로 하는 Population-based training이 효과 적임을 보였다[ref]. population-based training에 다양한 강화 학습 정책을 생성하는 'Trajectory Diversity (TrajeDi)' 기법을 적용하여 협력 게임에서 성능을 향상 시켰다[ref]. Self-play를 통해 얻은 agent는 unseen한 파트너를 만날 시 distributional shift로 인한 성능 저하가 발생하므로, 이를 완화하기 위한 Maximum Entropy Population-based training를 제안.[ref] 본 논문은 Zero-shot Human-AI Coordination문제를 handle하기 위해서 서로 다른 환경에서 학습된 agent들로 구성된 population pool을 구성하고, 이를 이용하여 협력을 목적으로 하는 agent를 학습 시켰습니다.
% \input{2.Related_work/Leveraging Environmental Distance for Diversity Enhancement}

\section{Automatic Curriculum Design For Zero-Shot Human-AI Coordination}
\label{sec:Method}
\subsection{Overview}

\begin{algorithm}[ht]
\caption{Automatic Curriculum Design for Zero-Shot Human-AI Coordination}
\label{alg:method}
\textbf{Input:} Training environments $\Lambda_{train}$ \\
\textbf{Initialise:} Ego policy $\pi$, Co-player population $\mathfrak{B}$ \\
\textbf{Initialise:} Co-player Env buffers $\forall \pi' \in \mathfrak{B}, \bm{\Lambda}(\pi') := \emptyset$ \\
\vspace{-\baselineskip}
\begin{algorithmic}[1]
\FOR {$i = \{1,2, \dots\}$}
    \FOR {$N$ episodes}
    \STATE $\pi' \sim \mathfrak{B}$ \algcom Sample co-player via Eq.\ref{eq:player_sampling}
    \STATE Sample \textit{Replay-decision}
    \IF {replaying}
        \STATE $\theta \sim \bm{\Lambda}(\pi')$ \algcom Sample a replay env with dist.\ref{eq:replay_dist} 
        \STATE Collect trajectory $\tau$ of $\pi$ using $(\theta, \pi')$
        \STATE Update $\pi$ with rewards $\bm{R}$ $(\tau) $
    \ELSE
        \STATE $\theta \sim \Lambda_{unseen/train}$ \algcom Sample unseen random env
        \STATE Collect trajectory $\tau$ of $\pi$ using $(\theta, \pi')$ 
    \ENDIF
        \STATE Compute Coordination score $S = {Return(\theta,\pi')}$   
        \STATE Update $\bm{\Lambda}(\pi')$ with $\theta$ using score $S$  
    \ENDFOR
    \STATE $\mathfrak{B} \gets \mathfrak{B} \cup \{\pi_i^{\perp}\}$, $\bm{\Lambda}(\pi_i^{\perp})$ := $\emptyset$ \algcom frozen weights
\ENDFOR
\end{algorithmic}
\end{algorithm}

In this section, we describe our proposed method, Automatic Curriculum Design for Zero-Shot Human-AI Coordination, which is an auto-curriculum that trains the ego-agent using return-based utility functions. Our method focuses on training the ego-agent to coordinate well with human partners in diverse environments. It is based on the MAESTRO algorithm \cite{samvelyan2023maestro}, which was developed for multi-agent UED. Our method extends MAESTRO to human-AI coordination settings. It curates environment/co-player pairs with learning potential for the ego-agent in human-AI coordination.

The training procedure of our proposed method is shown in Algorithm \ref{alg:method}. It begins by sampling the co-player agent $\pi'$ that coordinates with the ego-agent from the co-player population during training. This sampling is based on coordination performance (return), as described in Section \ref{sec:co-player sampling}. In the first epoch, the ego-agent coordinates with itself using the self-play method without sampling the co-player. 

After co-player sampling, our method employs a \textit{Replay-decision} based on a Bernoulli distribution. This decision process determines whether to sample an environment $\theta$ from the co-player's environment buffer or randomly sample an unseen environment $\theta$ from the finite set of training environments not yet stored in the co-player’s environment buffer. When an environment is sampled from the co-player's environment buffer, our method follows the replay distribution given by Equation (\ref{eq:replay_dist}), which is based on the return. It then collects trajectories from the ego-agent in the sampled environment/co-player pairs. The ego-agent policy is updated whenever an environment is sampled from the co-player's environment buffer. Coordination scores are calculated using the collected trajectories and updated in the individual environment buffer of the sampled co-player accordingly.
After N episodes of training, the ego agent's frozen weight policy is added to the co-player population $\mathfrak{B}$ with the environment buffer.

\begin{equation}
\label{eq:replay_dist}
\begin{aligned}
P_{\text{replay}} = (1-\rho) \cdot P_S + \rho \cdot P_C
\end{aligned}
\end{equation}

The replay distribution is a combination of two distributions: ($P_S$), based on the learning potential score, which is measured using the return. ($P_C$), based on how long each environment was sampled from the environment buffer. This equation follows the previous replay mechanism PLR \cite{jiang2021prioritized}.  The staleness coefficient $\rho \in [0, 1]$ is a weighting hyperparameter that balances ($P_S$) and ($P_C$). For more details on $P_S$, see Section \ref{sec:low_return}.

%%%%%%%%evaluation map %%%%%%%%%%
\begin{figure*}[!t]
  \centering
  \subfloat[$test_0$]{\includegraphics[width=0.19\linewidth]{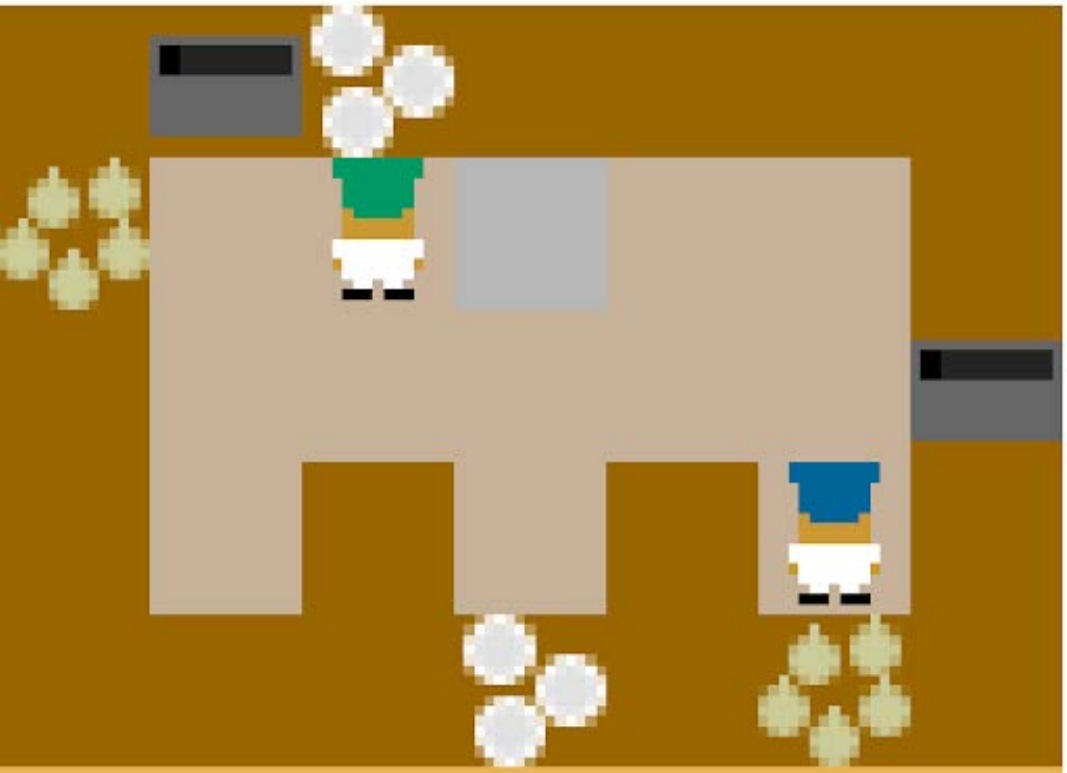}}
  \hspace{0.005\linewidth} % 그림 사이의 간격 조정
  \subfloat[$test_1$]{\includegraphics[width=0.19\linewidth]{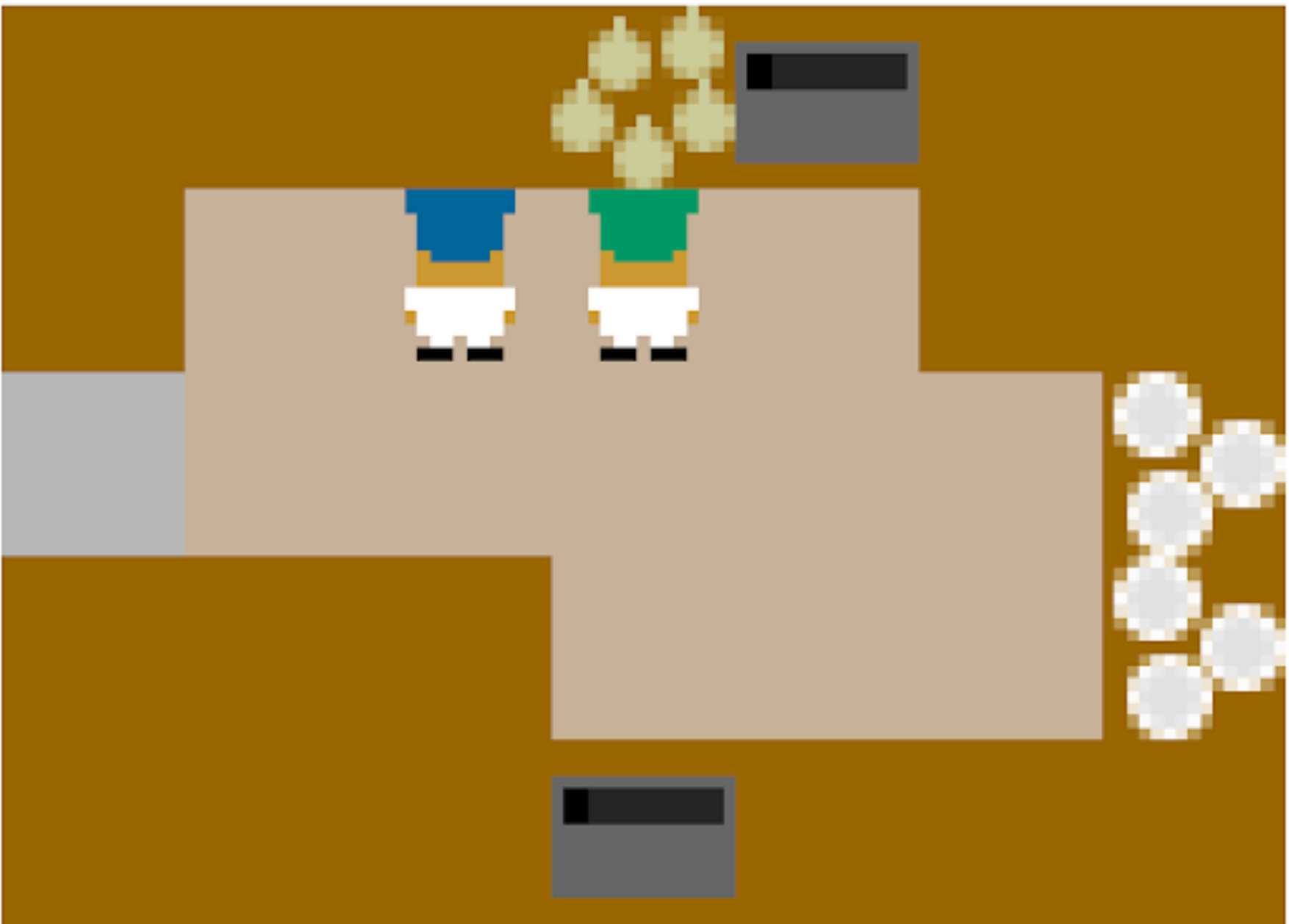}}
  \hspace{0.005\linewidth} % 그림 사이의 간격 조정
  \subfloat[$test_2$]{\includegraphics[width=0.19\linewidth]{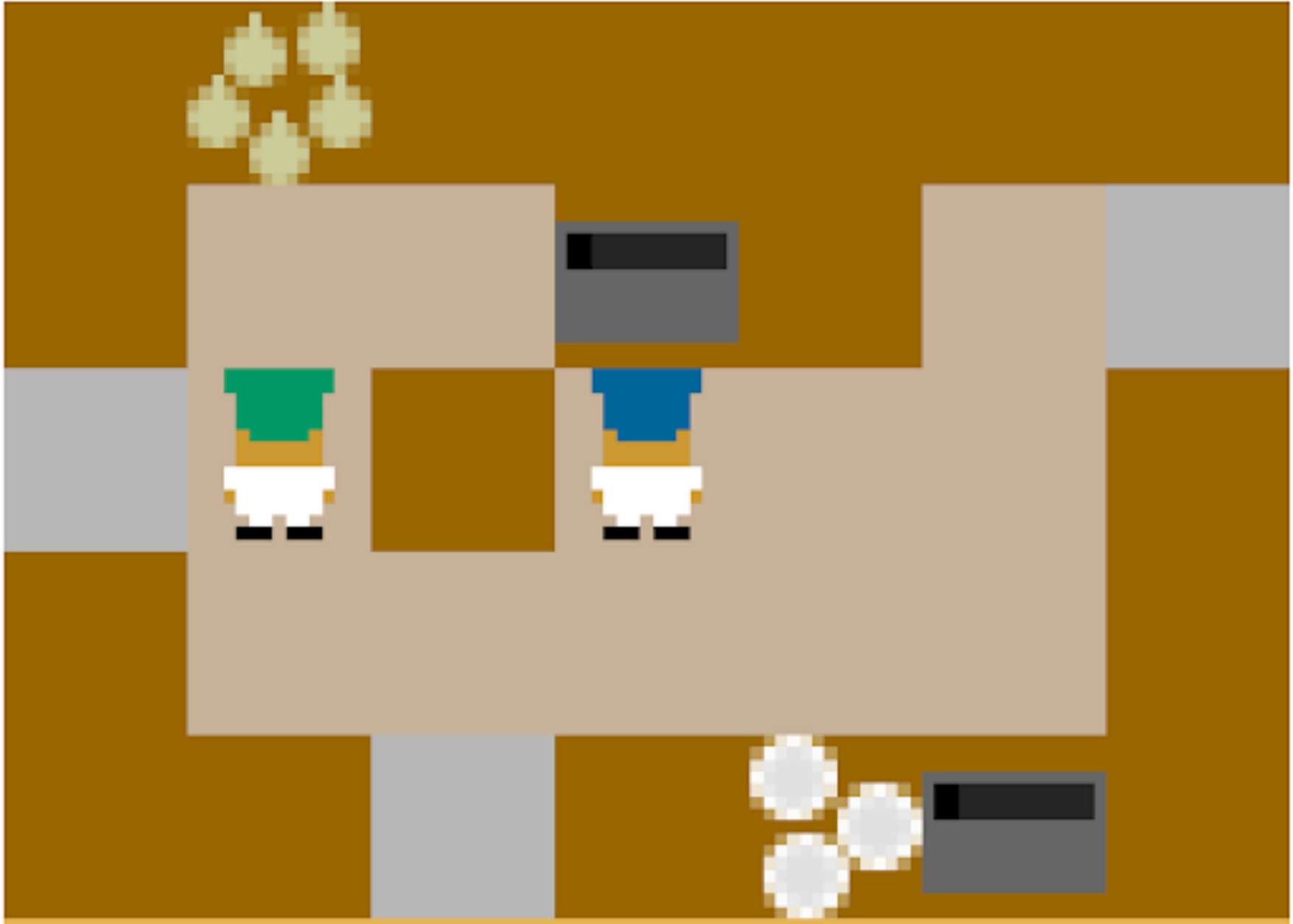}}
  \hspace{0.005\linewidth} % 그림 사이의 간격 조정
  \subfloat[$test_3$]{\includegraphics[width=0.19\linewidth]{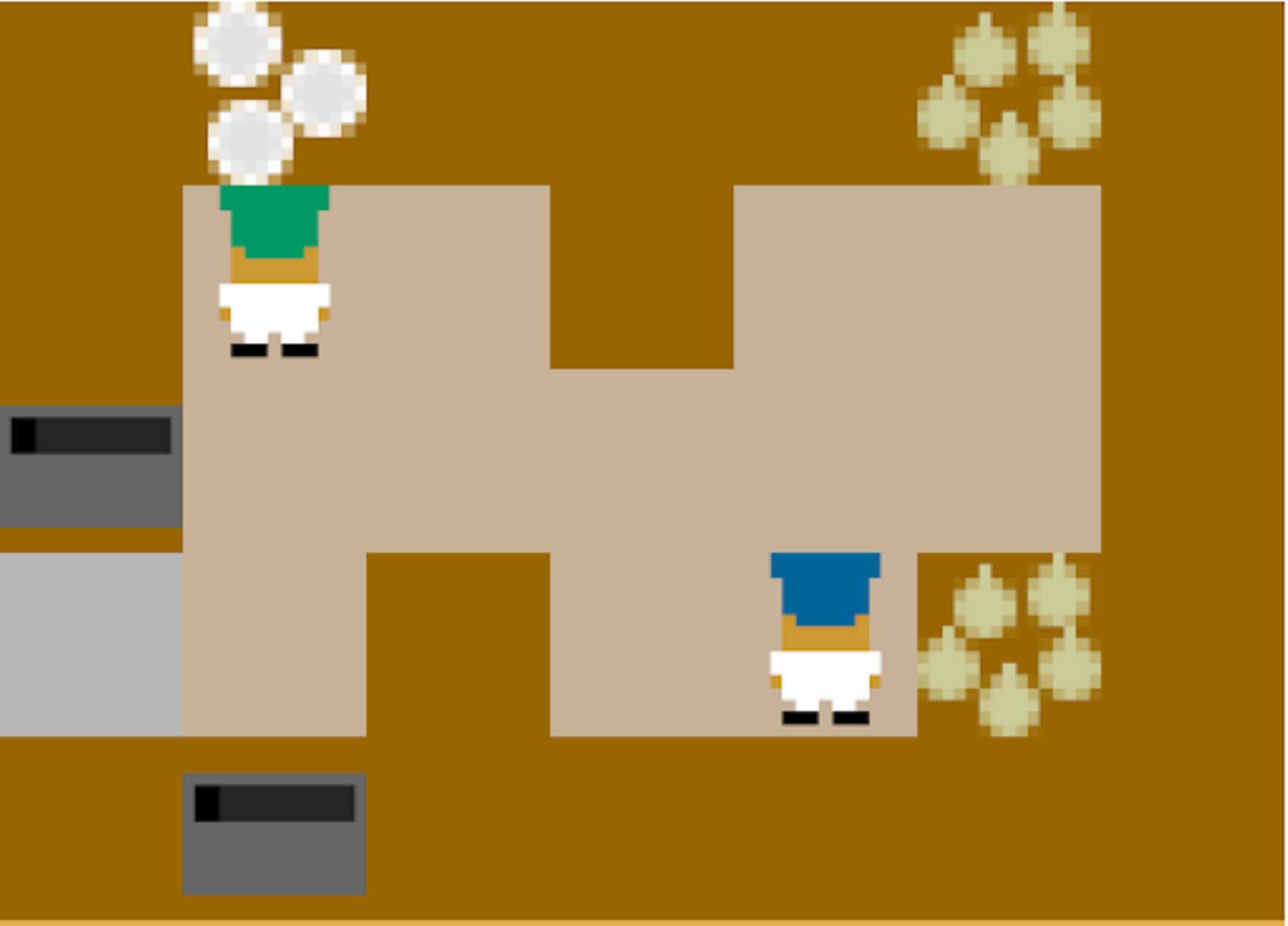}}
  \hspace{0.005\linewidth} % 그림 사이의 간격 조정
  \subfloat[$test_4$]{\includegraphics[width=0.19\linewidth]{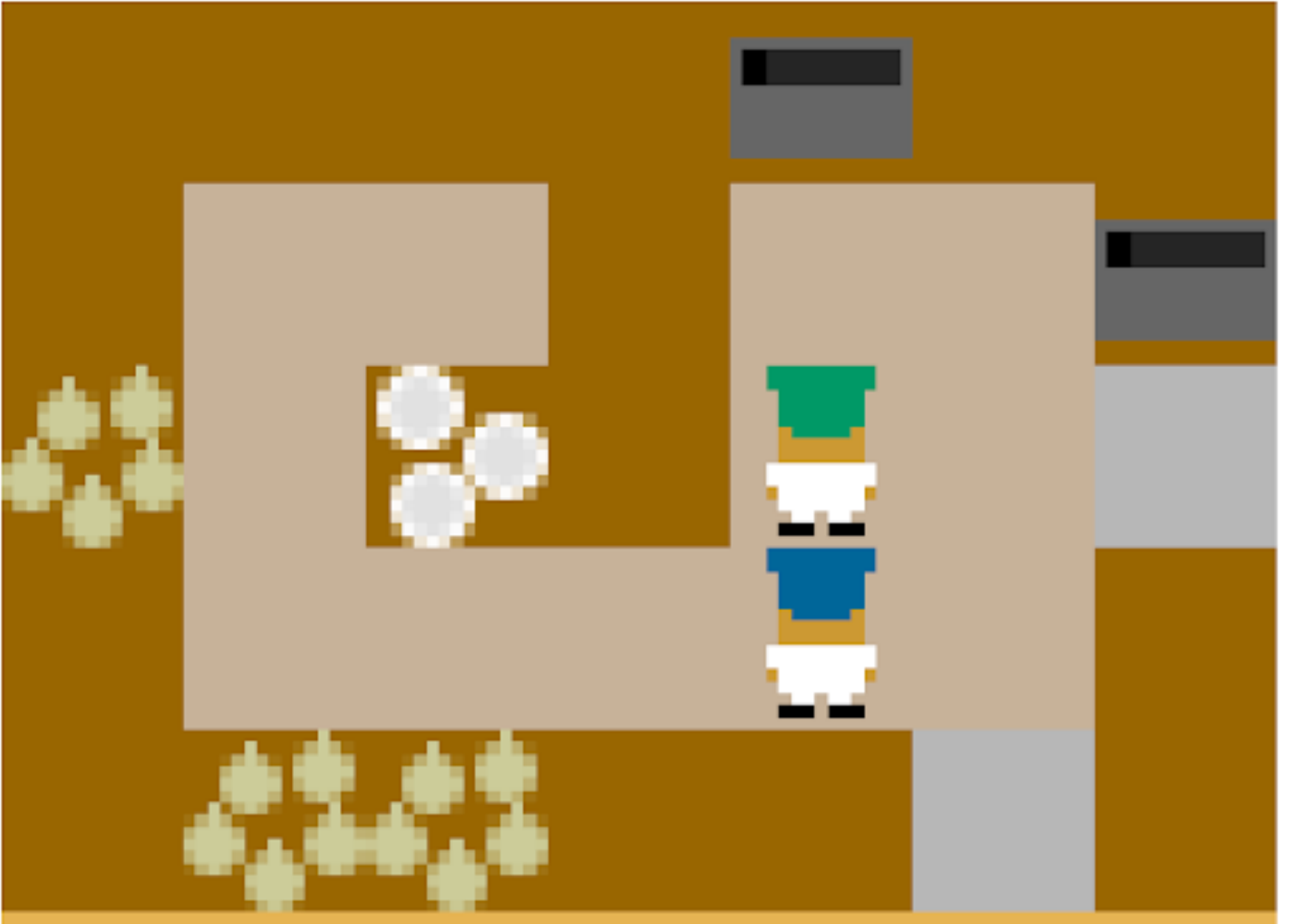}}
  \caption{\textbf{Evaluation Layouts.} The evaluation layouts were selected based on a previous study \protect \cite{yang2022optimal}. These layouts are selected to cover a range of difficulties, from easy ($test_{0}$) to difficult ($test_{4}$). Easier layouts allow agents to work independently within separate areas, whereas harder layouts require coordination with other agents without collisions.}
  \label{fig:eval-layouts}
\end{figure*}

\subsection{Prioritized Co-player Sampling}
\label{sec:co-player sampling}

The multi-agent UED study for two-player zero-sum games \cite{samvelyan2023maestro} prioritizes sampling the co-player with the highest regret environment in its buffer and focuses on minimizing regret to obtain a minimax regret policy. 

However, this approach is hard to apply directly to the human-AI coordination setting (common pay-off game). The ego-agent needs to learn a policy that not only minimizes regret but also considers the reward from coordination interaction, since the ego and co-player agent receive the common reward together. We need to set the co-player sampling metric by considering coordination for human-AI coordination. Most human-AI coordination studies \cite{zhao2023maximum,lou2023pecan} use return as a co-player sampling metric. They sample co-players who achieve low returns when coordinating with the ego-agent, which indicates challenging coordination scenarios. This approach optimizes the lower bound of coordination performance with any co-player in the population, ensuring that the ego-agent can coordinate well with various co-players and human players.

Unlike previous human-AI coordination approaches that sample co-players based on low returns in a static environment, our method jointly considers both the co-player and the environments they have encountered during training. Equation (\ref{eq:player_sampling}) formalizes this strategy by sampling the co-player whose buffer contains the lowest-return environment in the co-player population. Each co-player environment buffer has a fixed size of \textit{k} and stores environments based on the lowest return. It combines two existing ideas: (1) sampling co-players with low returns to improve worst-case coordination \cite{zhao2023maximum,lou2023pecan}, and (2) prioritizing high-potential learning experiences from environment buffers \cite{jiang2021prioritized}. 

\begin{equation} \label{eq:player_sampling}
    \textsc{Co-Player} \in \arg\min_{\pi' \in \mathfrak{B}} \left\{ \min_{\theta \in \bm{\Lambda}(\pi')} Return(\theta,\pi') \right \}
\end{equation}

Here, $\mathfrak{B}$ is the co-player population, consisting of ego-agent policies that are periodically frozen during training, $\bm{\Lambda}(\pi')$ is the environment buffer of the co-player agent $\pi'$, and $Return(\theta,\pi')$ is the return of the environment/co-player pairs $(\theta, \pi')$. 

\subsection{Learning Potential Score Based on Return}\label{sec:low_return}
Our method follows the replay distribution (Equation \ref{eq:replay_dist}) when selecting an environment for the ego-agent from an individual co-player buffer. In this replay distribution, $P_S$ is the first priority component that assigns higher priority to environments with learning potential for training the ego-agent. 

As discussed in Section \ref{sec:co-player sampling}, return is the appropriate scoring metric to build a prioritization distribution in human-AI coordination. It represents the coordination performance during the episode. Previous works \cite{zhao2023maximum,lou2023pecan} have demonstrated that return captures coordination performance and also indicates which environments require more training. If the replayed environment has a low return, it indicates that the ego-agent is hard to coordinate in that environment. The details of $S$ are as follows: 

\begin{equation} \label{eq:return}
    S_{i} = \frac{1}{T} \sum_{t=0}^{T} \text{Return}_t(\theta_{i}, \pi')
\end{equation}

\begin{equation} \label{eq:potential_score}
    P_{S}(\theta_{i}\vert S) = \frac{\textrm{rank}\left({S_{i}}\right)^{\frac{1}{\beta}}}{\sum_{j} \textrm{rank} \left({S_{j}}\right)^{\frac{1}{\beta}}}
\end{equation}

Equation (\ref{eq:return}) is the expected return of the replayed environment with the sampled co-player during the episode. Here, $T$ represents the total number of timesteps in the episode, $t \in \{0, \dots, T\}$ denotes each timestep within the episode, and $i$ denotes the index of the replayed environment from the co-player's buffer. This environment score is normalized using Equation (\ref{eq:potential_score}). This prioritization approach adapts the temperature-controlled ranking of previous work \cite{jiang2021prioritized} to our coordination setting. We normalize the scores using a rank-based prioritization, which sorts the scores in the co-player buffer in ascending order, prioritizing low-return environments. $j$ denotes the index of environments in the co-player's buffer. $\beta\in [0, 1]$ is the temperature parameter that adjusts the influence of the distribution.

\section{Experiments Setup}
\label{sec:Setup}
%%%%%%%% result %%%%%%%%%%            

Overcooked AI \cite{carroll2019utility} is a cooperative game played in a grid-world environment inspired by the kitchen. The goal is to cook as much food as possible in a limited time. Agents can get rewards by completing tasks sequentially, such as preparing ingredients, cooking, and serving dishes. Cooperation is essential in this environment, where success depends on effective coordination between players and their partners.

\noindent\textbf{Baseline Model.} We compare our method with three key baseline models related to UED: MAESTRO \cite{samvelyan2023maestro}, Robust PLR \cite{jiang2021prioritized}, and Domain Randomization \cite{jakobi1997evolutionary}. Robust PLR and domain randomization are methods originally used in a single-agent setting, so we modified them for multi-agent settings using the self-play method. We used the PPO method \cite{schulman2017proximal} to train the ego-agent. PPO is the standard algorithm for training ego-agents through population-based learning with self-play in zero-shot human-AI coordination research. Previous studies \cite{strouse2021collaborating} have demonstrated that this approach enables agents to coordinate with humans without relying on human data. The model architecture and hyperparameter choices are provided in Appendix \ref{sec_params}.

\noindent\textbf{Layout Generation.} In our experiments, we automatically generated a total of 6,000 different layouts for Overcooked-AI, each layout was 7 × 5. These layouts were generated using a heuristic-driven layout generator capable of adjusting three layout parameters: the number of interactive blocks (such as pots, plates, outlets, and onions), the number of spaces, and the size of the layouts, which is fixed in this study. The generator was designed to ensure that each layout included at least 12 spaces and 5 interactive blocks. To avoid redundancy, the layout generator compared a layout array during generation. In addition, we check the solvability of the layout using the A* path-finding algorithm \cite{hart1968formal}, which can find a path for the agent to access any interactive block. Further details on the algorithm are provided in the Appendix \ref{sec_layout_generator}.

\noindent\textbf{Human Proxy.} To evaluate the coordination performance of the trained agent in our experiments, we used a human proxy agent from a previous study \cite{yang2022optimal} as a co-player agent. This human proxy agent performs as human-like as possible by adjusting its parameters based on human data using cross-entropy methods.

\noindent\textbf{Evaluation.}
We evaluated the human-AI zero-shot coordination performance on five unseen layouts (Figure \ref{fig:eval-layouts}), selected based on coordination difficulty. Specifically, the five layouts correspond to the 100th, 75th, 50th, 25th, and 0th percentiles of the maximum achievable self-play rewards, as in a previous study \cite{yang2022optimal}. This helps to measure the adaptability of the ego-agent based on the difficulty of the layout. In addition, we conducted a utility function analysis.

\begin{figure*}[ht]
  \centering
  \subfloat[\textbf{Overall Results}]{\includegraphics[width=0.48\linewidth]{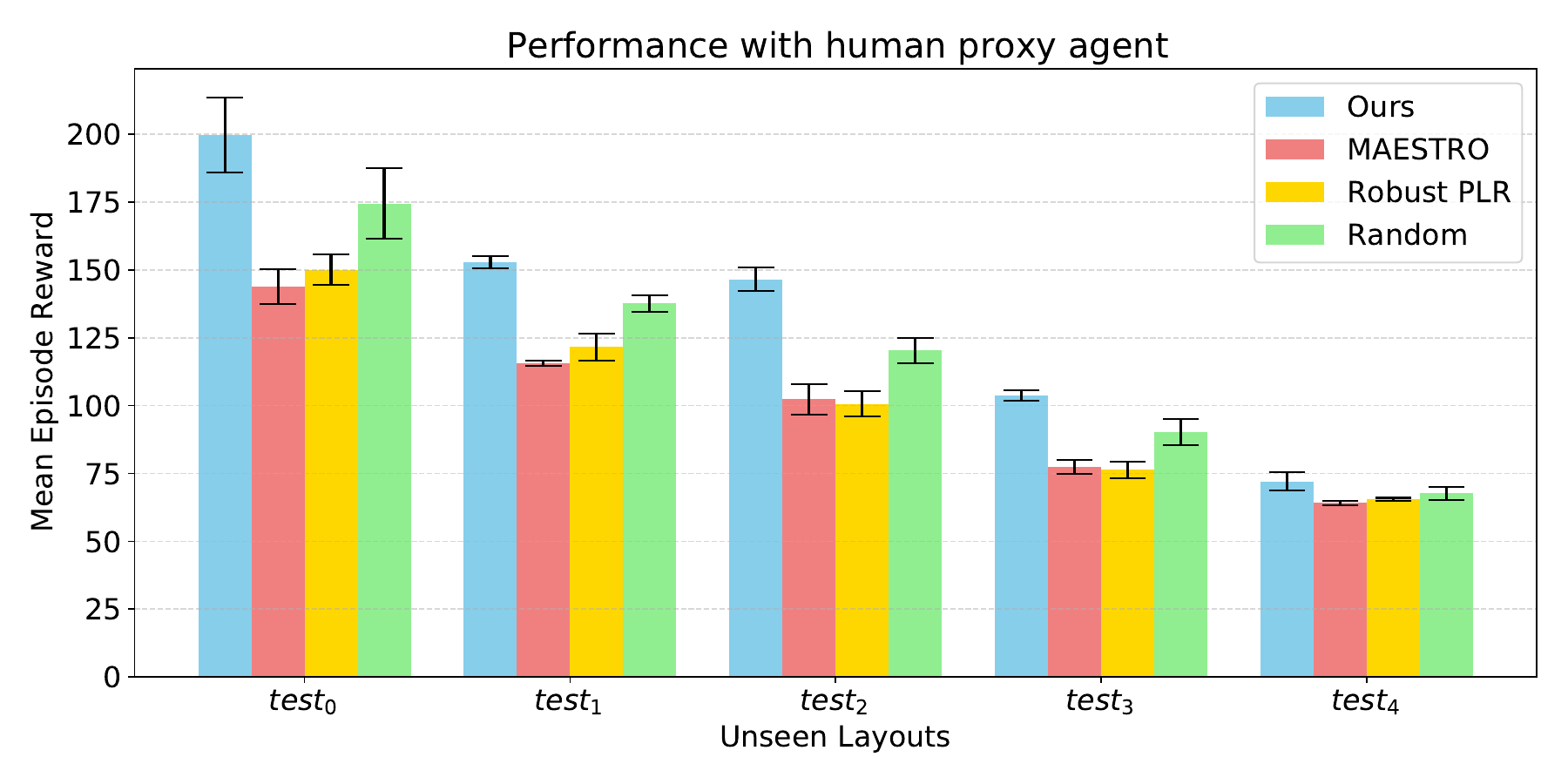}}
  \subfloat[\textbf{Utility Function Analysis}]{\includegraphics[width=0.48\linewidth]{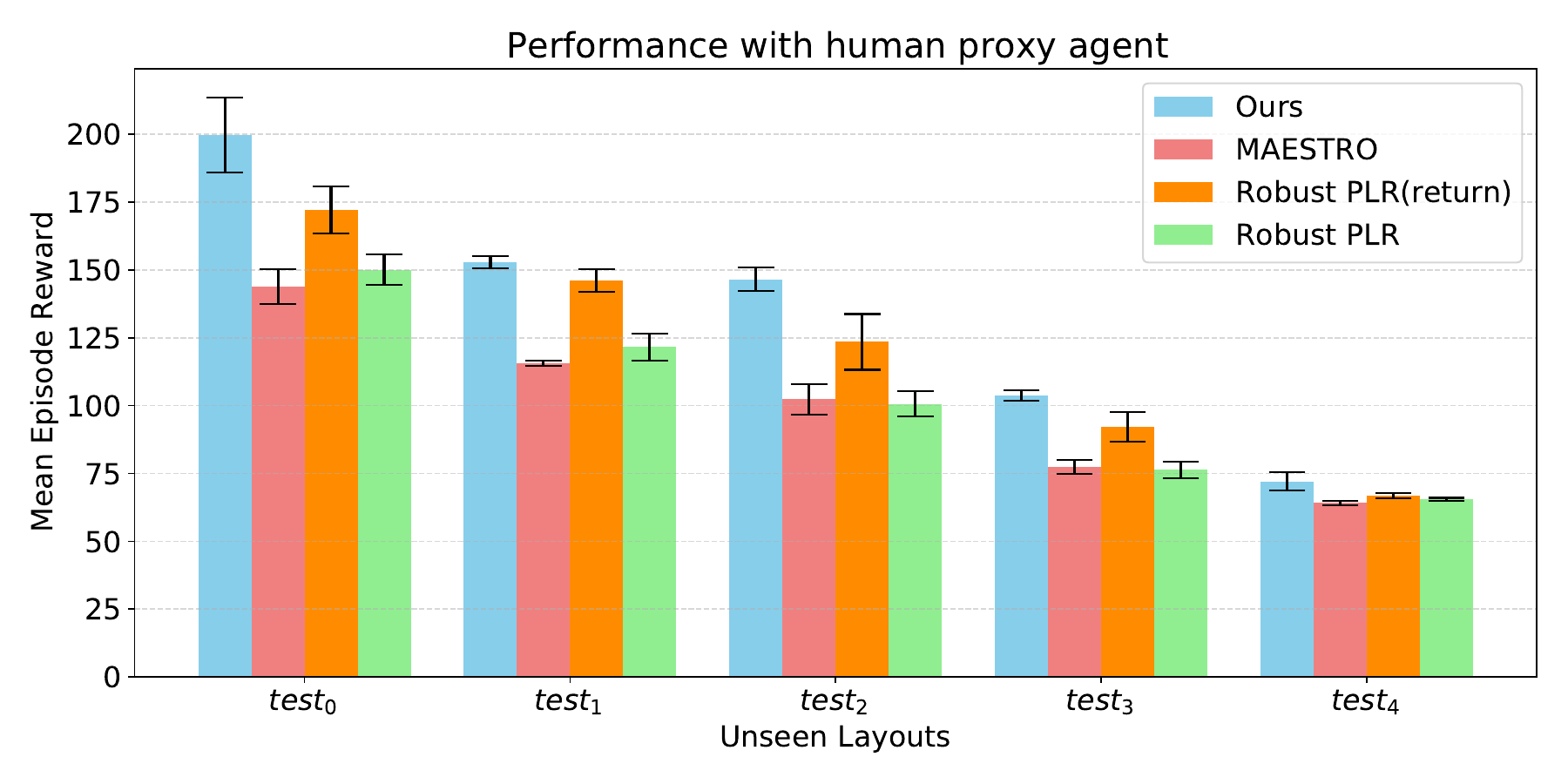}}
   \caption{ \textbf{(a) Overall results with a human proxy model.} Our method achieves higher coordination performance with the human proxy agent than other baselines on all five evaluation layouts (Mean and standard error were computed over 100 runs with three random seeds). \textbf{(b) Utility function analysis ($Regret$ vs $Return$).} Using a return-based utility function has an advantage over a regret-based utility function in training an ego-agent for the regret-based utility function for multi-agent coordination settings (Mean and standard error were computed over 100 runs with three random seeds).}
  \label{main_res}
  % \vspace{-10pt}
\end{figure*}

\vspace{-1em}

\begin{figure*}[ht]
    \subfloat[\textbf{Training}]
    {\includegraphics[width=0.33\textwidth]{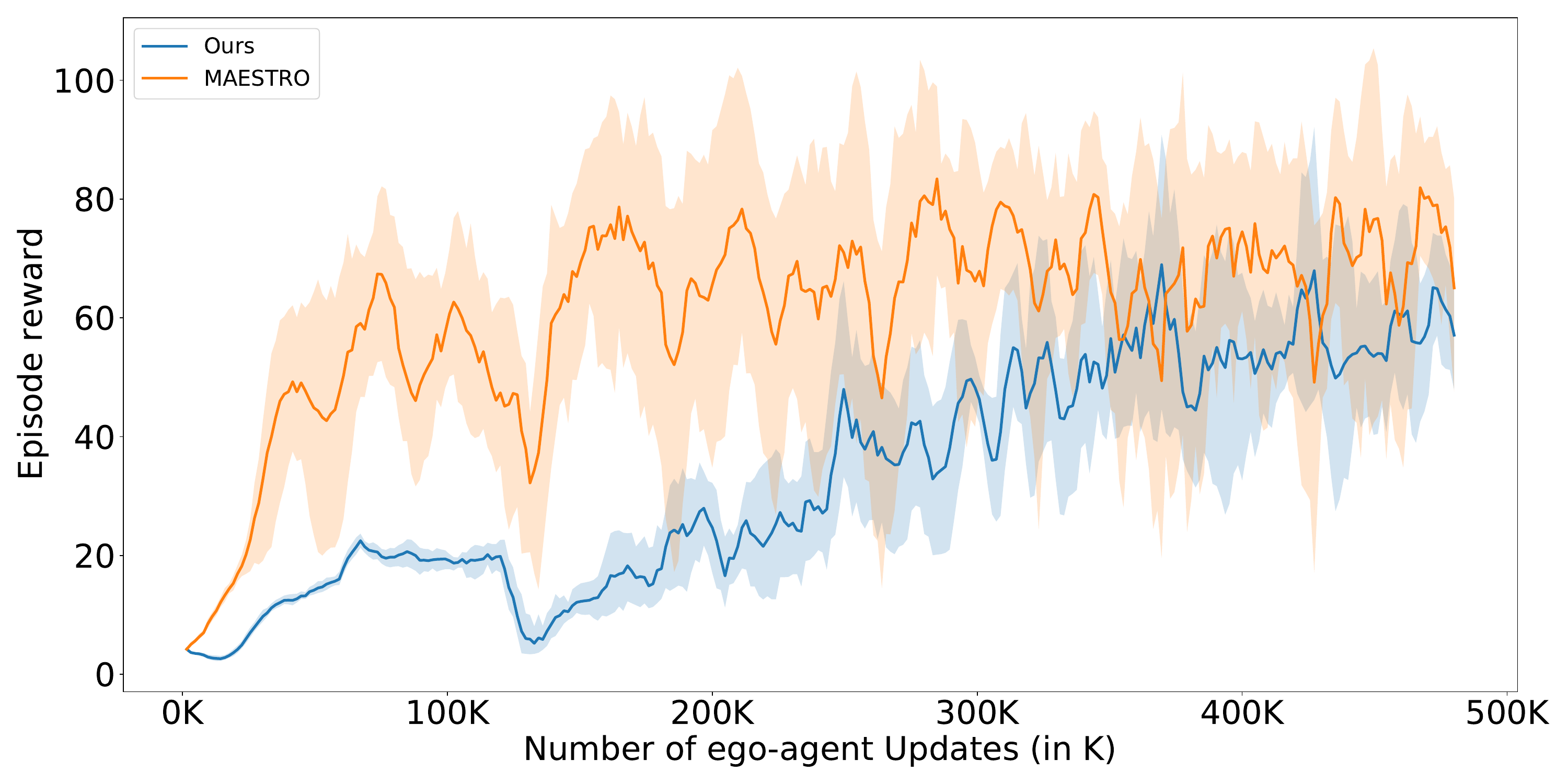}}
    \subfloat[\textbf{Training-evaluation}]{\includegraphics[width=0.33\textwidth]{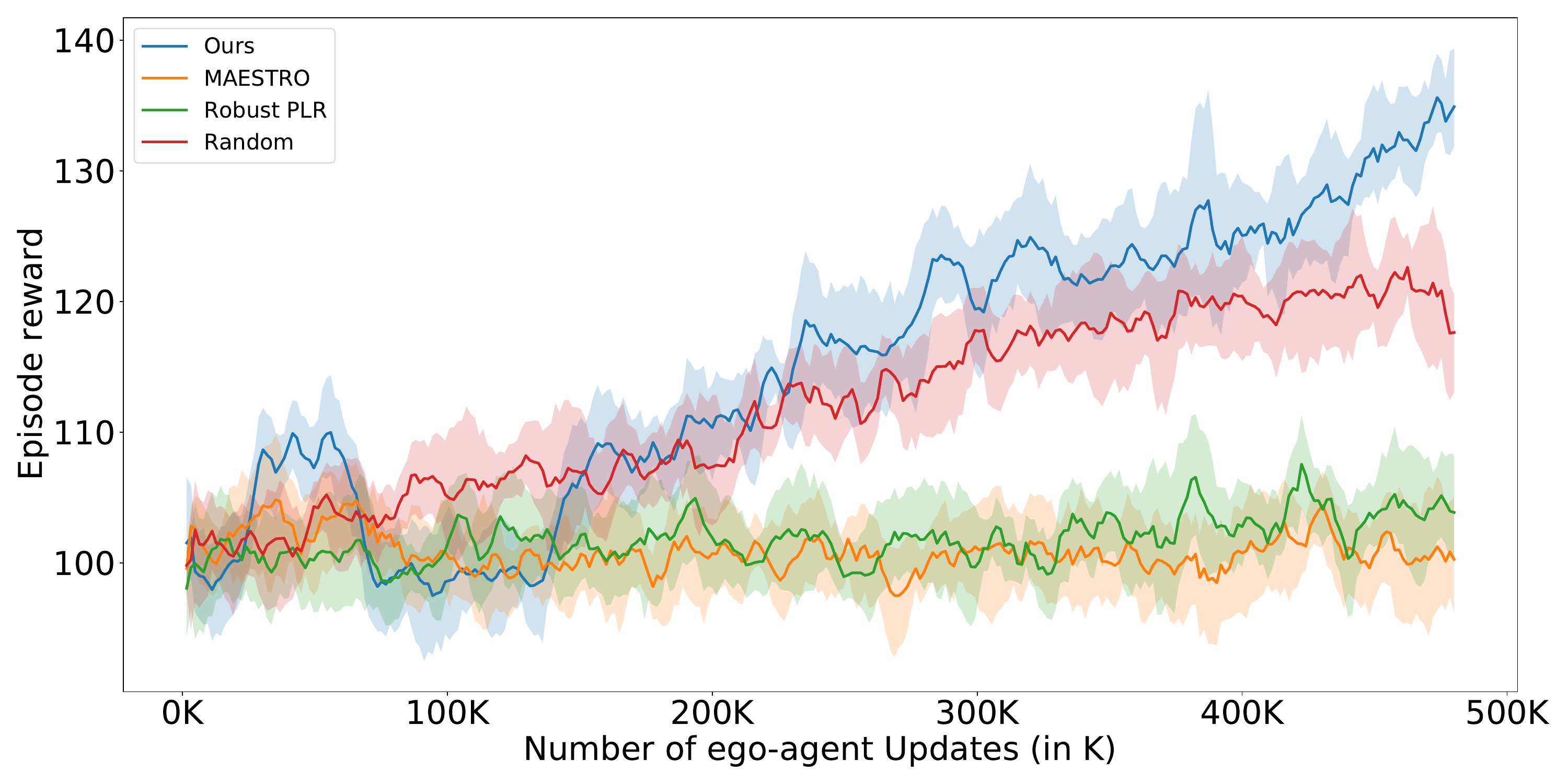}}
    \subfloat[\textbf{Evaluation}]
    {\includegraphics[width=0.33\textwidth]{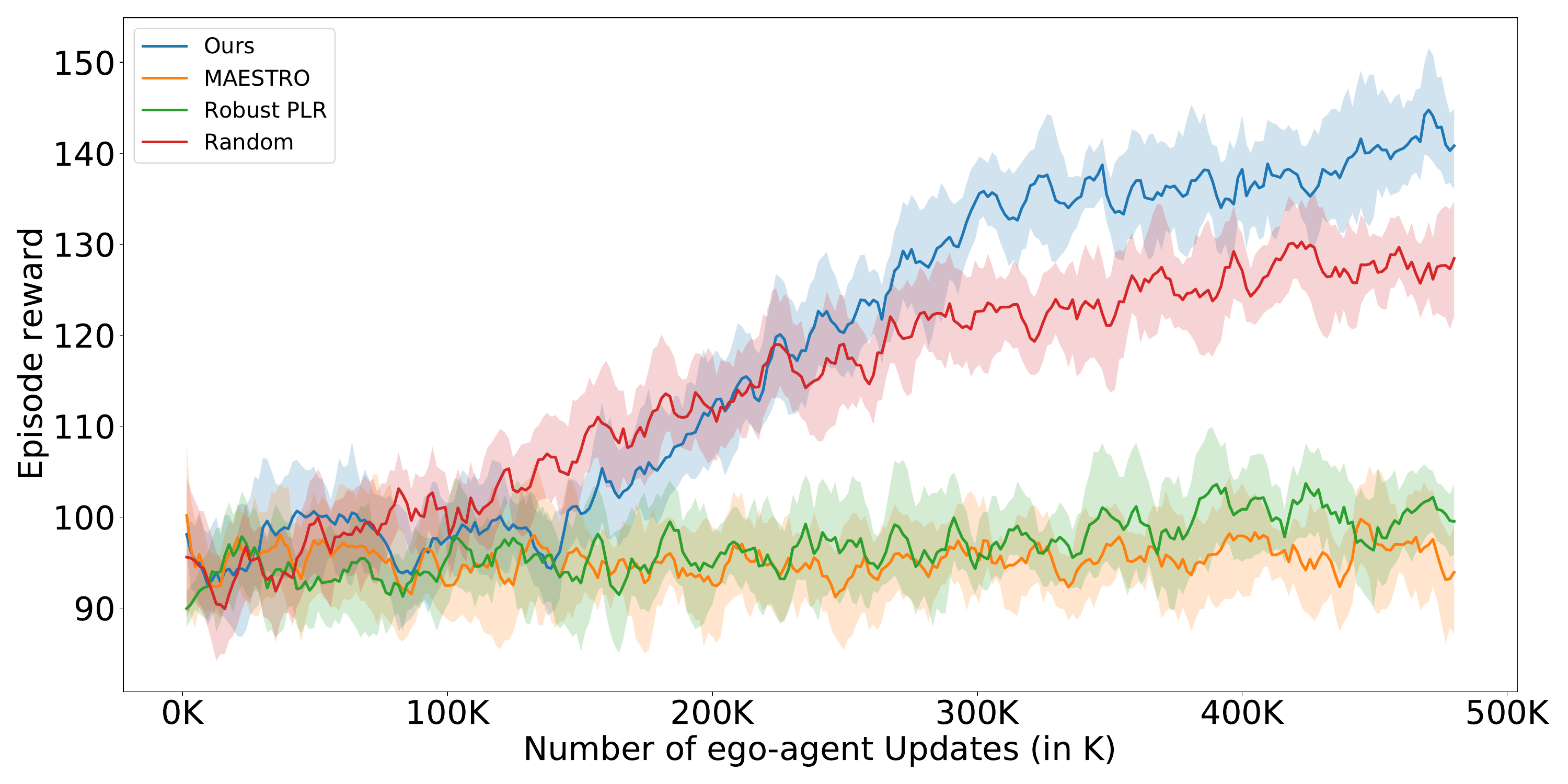}}
    \caption{ \textbf{Learning curves for the training, training-evaluation, and evaluation phases.} The solid line shows the average episode reward across the three different random seeds, while the shaded areas show the maximum and minimum rewards.  The results show that our model achieves higher performance on both the training and evaluation layouts.}
    \vspace{-0.5em}
    \label{training_graph}
\end{figure*}

\begin{figure}[ht]
    {\includegraphics[width=0.5\textwidth]{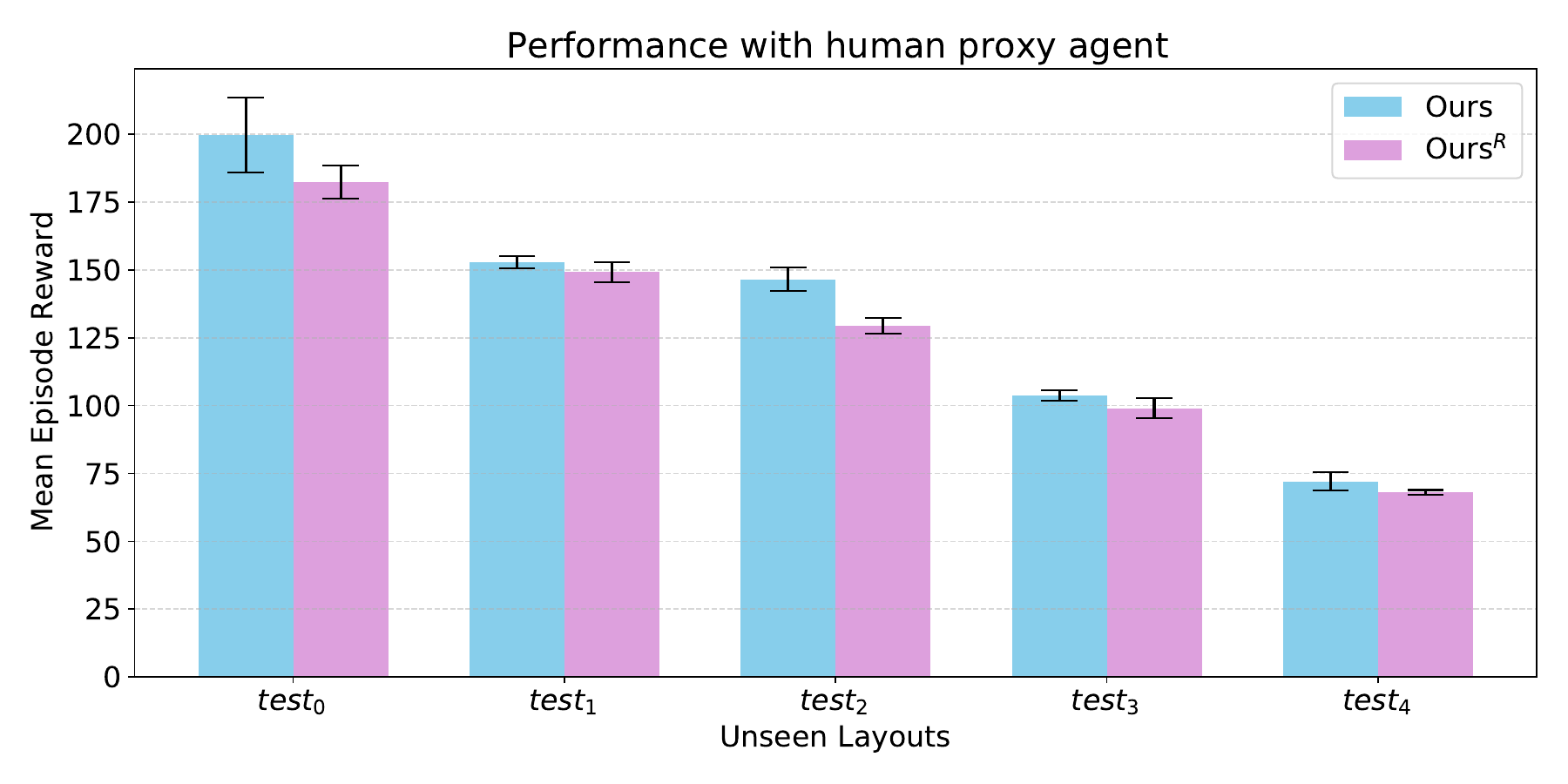}}
    \caption{\textbf{Results of the prioritized co-player sampling} Prioritized co-player sampling leads to higher coordination performance with the human proxy agent compared to random co-player sampling across all five evaluation layouts. (Mean and standard error were computed over 100 runs with three random seeds.) Ours$^{R}$ represents our proposed method with random co-player sampling.}
    \label{sampling_graph}
    \vspace{-1em}
\end{figure}

 \begin{figure*}[ht]
    \centering
    \subfloat[\textbf{Human Study Result.}]{\includegraphics[width=0.48\textwidth]{
    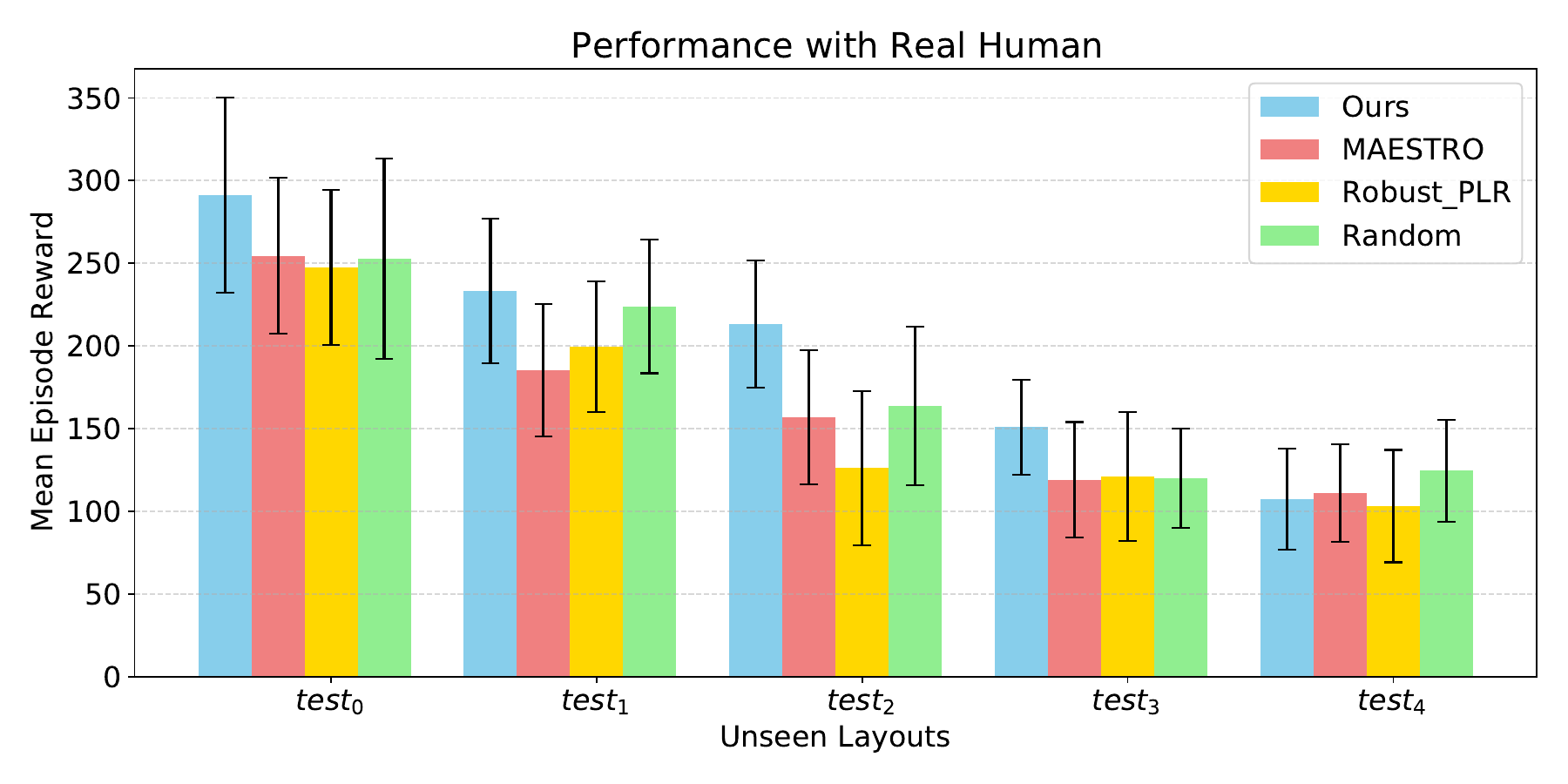}}
    \subfloat[\textbf{Human Survey Result.}]{\includegraphics[width=0.48\textwidth]{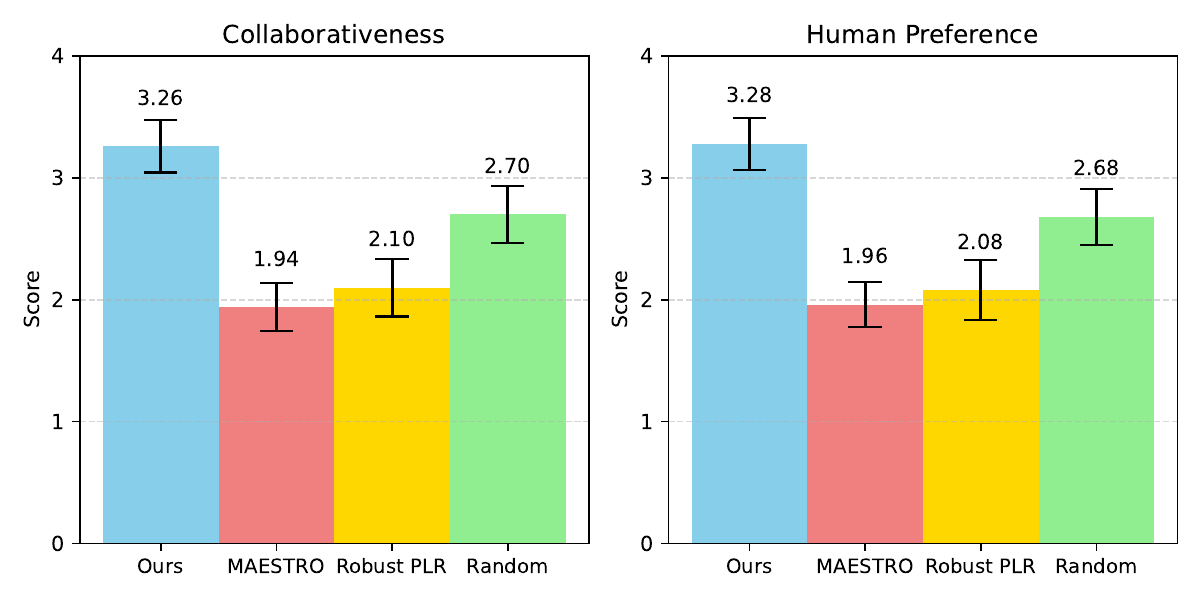}}
    \caption{This plot shows the results of an experiment with 20 real humans: (a) The average reward obtained on the five unseen evaluation layouts, (b) The average evaluation scores for the two items "Collaborativeness" and "Human Preference". Each subject was scored on a scale of 1 to 4 based on their voting ranking. A higher score indicates a more positive evaluation.}
    \vspace{-0.5em}
    \label{human-ai}
\end{figure*}

\section{Experiments Result}
\label{sec:Result}
\subsection{Experiments With Human Proxy}
\noindent\textbf{Overall Result.} Figure \ref{main_res}(a) shows the comparison result of our method and the baseline models (MAESTRO, Robust PLR, and Domain Randomization) in terms of coordination performance with a human proxy agent on unseen layouts. The bar graph shows the evaluation results, which are the average episode rewards (three random seeds) with the human proxy agent on the evaluation layouts. As a result, our method showed better performance in all five evaluation layouts compared to other baselines. This performance gap is statistically significant across all layouts except $test_{4}$, as shown by the results of the paired t-test in Appendix~\ref{sec_t_test} (see Table \ref{tab:t_test_human_proxy}). We observe that our method achieves a higher generalization coordination performance with unseen partners on unseen evaluation layouts. We also performed cross-play experiments, where our method outperformed other baselines in unseen evaluation layouts. The results of these experiments are provided in the Appendix \ref{sec_crossplay}.

\subsection{Study of the Utility Function}

\noindent\textbf{Evaluation Result.} We conducted an analysis to try to figure out which utility functions are appropriate for a human-AI coordination setting. Figure \ref{main_res}(b) shows how the utility function affects the training of the ego-agent. Our method and Robust PLR (return) outperformed MAESTRO (regret-based baseline) and Robust PLR in all five evaluation layouts with the human proxy agent, demonstrating the effectiveness of return over regret as a utility function.

In this experiment, return is a more appropriate utility function than regret in human-AI coordination settings. The regret utility function is not suitable for curating challenging environments for coordination tasks. Regret quantifies the opportunity cost of not making the optimal decision without considering coordination. Therefore, it is difficult to assess the effectiveness of coordination performance using regret. However, return is appropriate for evaluating coordination performance because it is derived from the common rewards that agents receive through their collaborative efforts to achieve a common goal.

\noindent\textbf{Learning Curve.} Figure \ref{training_graph}(a) shows the learning curves of our method and MAESTRO during training. By comparing these learning curves, we evaluate which utility function is more effective for training the ego-agent in human-AI coordination. MAESTRO shows high episode reward in the early training stage due to overfitting easy difficulty environments, but shows high fluctuation when it encounters relatively less seen environments. On the other hand, our method steadily increases episode rewards by sampling more difficult environments according to the agent's coordination ability. Appendix \ref{sec_buffer} shows how the buffer composition of each method changes as the training progresses. Figure \ref{training_graph}(b) illustrates the coordination performance of the ego-agent with the human proxy agent on randomly selected layouts from the training set at each training step. Meanwhile, Figure \ref{training_graph}(c) shows the evaluation graph of the coordination performance of the ego agent with the human proxy agent in the evaluation layouts. The use of return as a utility function leads to an increase in episode rewards, whereas regret does not. This suggests that the ego-agent trained with a return-based approach achieves more effective coordination with the human proxy agent.

\subsection{Study of the Co-player Sampling}

\noindent\textbf{Evaluation Result.} To evaluate the effect of prioritized co-player sampling, we compared our proposed method with random co-player sampling.
Figure \ref{sampling_graph} shows that prioritized co-player sampling improves the training performance of the ego-agent. Our sampling strategy prioritizes co-players with low returns and jointly considers the challenging environments they have encountered. As a result, the ego-agent optimizes the lower bound of coordination performance. In contrast, random co-player sampling does not account for the varying difficulty levels of co-players or environments. This comparison highlights the importance of curating challenging coordination scenarios for the ego-agent.

\subsection{Human-AI coordination}

\noindent\textbf{Human Study Result.} Additionally, we conducted a human study to evaluate coordination with real humans. We recruited 20 participants from the local student community and evaluated the proposed method and the baseline models with our participants. All participants provided their informed consent before the experiment, which was conducted following ethical guidelines approved by the Institutional Review Board (IRB). 

Figure \ref{human-ai} shows the average coordination performance of five evaluation layouts in the human study. Our method outperformed the baseline models on all layouts ($test_{0}$, $test_{1}$, $test_{2}$, $test_{3}$), but showed a similar performance to other baselines in $test_{4}$, with no statistically significant differences ($p > 0.05$) according to a t-test. This performance is a limitation of our approach. Our method is designed to improve the lower bound of coordination performance by prioritizing training with difficult environment/co-player pairs, which enhances robustness in most scenarios. However, this strategy may bias the model towards specific coordination patterns, potentially limiting generalization in extremely complex environments like $test_{4}$. The sophisticated coordination requirements of $test_{4}$ may fall outside the range of patterns for which our method has been optimized. However, consistent performance improvements in $test_{0}$ through $test_{3}$ support the effectiveness of our method. The detailed results of the paired t-tests for all layouts are provided in Appendix \ref{sec_t_test} (see Table \ref{tab:t_test_human}).

\noindent\textbf{Human Survey Result.} For each session of the experiment, the participants ranked the models in terms of collaborativeness (the most collaborative partner)  and human preference (their top preference). The human evaluation scores in Figure \ref{human-ai} (b) were converted to a scale of 1 to 4 based on each participant's subjective ranking of the models. A higher score indicates a better evaluation. The plot shows that our method has higher collaborativeness and human preference compared to the baseline models. Participants rated the agent trained with our method as more cooperative and preferred it for coordination over baseline models. Collaborativeness and human preference scores for each layout are provided in the appendix \ref{sec_buffer}.

\section{Conclusion}
\label{sec:Conclusion}

\noindent\textbf{Summary.} 
In this paper, we propose Automatic Curriculum Design for Zero-Shot
Human-AI coordination. Our method extends the multi-agent Unsupervised Environment Design (UED) approach to zero-shot human-AI coordination, which trains the ego-agent to coordinate with human partners in unseen environments. We use return as a utility function, unlike previous multi-agent UED methods that use regret. Our method outperformed other baselines when evaluated with a human proxy in Overcooked-AI. Ablation studies demonstrate the effectiveness of the proposed components. Finally, we had a real human experiment to evaluate the human-AI coordination performance of our method. The result of the human-AI coordination experiment shows that our method outperforms other baselines.

\noindent\textbf{Limitation and Future Work.}
When using the return as a measure of the learning potential of the environment/co-player pair, the evaluation performance improved compared to a previous method. However, prioritizing environments with low returns leads to the issue of sampling environment/co-player pairs that may be too challenging for the agent to learn effectively. In future work, we plan to adopt an alternative metric based on success probability \cite{rutherford2025no} to promote the sampling of pairs of learnable environment-co-players for human-AI coordination. The success of coordination can be evaluated through indicators such as collision-free movements and cooperative completion of cooking tasks. We will also use mutation to adjust environments that are sampled repeatedly to prevent overfitting to specific environments \cite{parker2022evolving,wang2020enhanced}. 

In the current multi-agent UED, the co-player pool is constructed from the past frozen weights of trained ego-agents. This approach mitigates the non-stationarity problem compared to self-play approaches, in which the co-player policy is changed during training. However, since it essentially plays with past versions of itself, it struggles to train diverse strategies. We plan to investigate methods for increasing the diversity of the co-player pool in UED.

%추가적인 limitation 혹은 future work?
%% 리턴기반의 방법은 

%In this paper, We develop an effective level sampling framework for generalization in  multi-agent learning. By formalizing the approach utilized in previous curriculum-related studies as a score function and integrating the dissimilarity metric to encourage the selection of diverse map configurations, Our method effectively enhances the generalization performance of the agent. While regret-based sampling, which has been widely adopted in prior research, exhibits diminished performance in cooperation environments, return-based sampling demonstrates  superior zero-shot performance compared to random sampling. Nonetheless, our work presents several promising avenues for future investigation. Firstly, the 
% proposed method requires validation across various environments and the exploration of alternative approaches to determine similarities between environments. Additionally, since the generalization problem in a multi-agent setting also necessitates considering the policies of co players, further research is warranted to  address the generalization of diverse agent populations.

% \section*{Acknowledgments}
% This work was supported by Institute of Information & communications Technology Planning & Evaluation (IITP) grant funded by the Korea government (MSIT) (No.2019-0-01842, Artificial Intelligence Graduate School Program (GIST))
% This research was supported the National Research Foundation of Korea(NRF) funded by the MSIT(2021R1A4A1030075)

%% The file named.bst is a bibliography style file for BibTeX 0.99c
\bibliographystyle{named}
\bibliography{ijcai25}

\begin{thebibliography}{}

\bibitem[\protect\citeauthoryear{Berner \bgroup \em et al.\egroup }{2019}]{berner2019dota}
Christopher Berner, Greg Brockman, Brooke Chan, Vicki Cheung, Przemys{\l}aw D{\k{e}}biak, Christy Dennison, David Farhi, Quirin Fischer, Shariq Hashme, Chris Hesse, et~al.
\newblock Dota 2 with large scale deep reinforcement learning.
\newblock {\em arXiv preprint arXiv:1912.06680}, 2019.

\bibitem[\protect\citeauthoryear{Carroll \bgroup \em et al.\egroup }{2019}]{carroll2019utility}
Micah Carroll, Rohin Shah, Mark~K Ho, Tom Griffiths, Sanjit Seshia, Pieter Abbeel, and Anca Dragan.
\newblock On the utility of learning about humans for human-ai coordination.
\newblock {\em Advances in neural information processing systems}, 32, 2019.

\bibitem[\protect\citeauthoryear{Chung \bgroup \em et al.\egroup }{2024}]{chungadversarial}
Hojun Chung, Junseo Lee, Minsoo Kim, Dohyeong Kim, and Songhwai Oh.
\newblock Adversarial environment design via regret-guided diffusion models.
\newblock In {\em The Thirty-eighth Annual Conference on Neural Information Processing Systems}, 2024.

\bibitem[\protect\citeauthoryear{Dennis \bgroup \em et al.\egroup }{2020}]{dennis2020emergent}
Michael Dennis, Natasha Jaques, Eugene Vinitsky, Alexandre Bayen, Stuart Russell, Andrew Critch, and Sergey Levine.
\newblock Emergent complexity and zero-shot transfer via unsupervised environment design.
\newblock {\em Advances in neural information processing systems}, 33:13049--13061, 2020.

\bibitem[\protect\citeauthoryear{Devlin \bgroup \em et al.\egroup }{2011}]{devlin2011empirical}
Sam Devlin, Daniel Kudenko, and Marek Grze{\'s}.
\newblock An empirical study of potential-based reward shaping and advice in complex, multi-agent systems.
\newblock {\em Advances in Complex Systems}, 14(02):251--278, 2011.

\bibitem[\protect\citeauthoryear{Fang \bgroup \em et al.\egroup }{2019}]{fang2019curriculum}
Meng Fang, Tianyi Zhou, Yali Du, Lei Han, and Zhengyou Zhang.
\newblock Curriculum-guided hindsight experience replay.
\newblock {\em Advances in neural information processing systems}, 32, 2019.

\bibitem[\protect\citeauthoryear{Guan \bgroup \em et al.\egroup }{2023}]{guan2023efficient}
Cong Guan, Lichao Zhang, Chunpeng Fan, Yichen Li, Feng Chen, Lihe Li, Yunjia Tian, Lei Yuan, and Yang Yu.
\newblock Efficient human-ai coordination via preparatory language-based convention.
\newblock {\em arXiv preprint arXiv:2311.00416}, 2023.

\bibitem[\protect\citeauthoryear{Gur \bgroup \em et al.\egroup }{2021}]{gur2021environment}
Izzeddin Gur, Natasha Jaques, Yingjie Miao, Jongwook Choi, Manoj Tiwari, Honglak Lee, and Aleksandra Faust.
\newblock Environment generation for zero-shot compositional reinforcement learning.
\newblock {\em Advances in Neural Information Processing Systems}, 34:4157--4169, 2021.

\bibitem[\protect\citeauthoryear{Hart \bgroup \em et al.\egroup }{1968}]{hart1968formal}
Peter~E Hart, Nils~J Nilsson, and Bertram Raphael.
\newblock A formal basis for the heuristic determination of minimum cost paths.
\newblock {\em IEEE transactions on Systems Science and Cybernetics}, 4(2):100--107, 1968.

\bibitem[\protect\citeauthoryear{Hu \bgroup \em et al.\egroup }{2020}]{hu2020other}
Hengyuan Hu, Adam Lerer, Alex Peysakhovich, and Jakob Foerster.
\newblock “other-play” for zero-shot coordination.
\newblock In {\em International Conference on Machine Learning}, pages 4399--4410. PMLR, 2020.

\bibitem[\protect\citeauthoryear{Jakobi}{1997}]{jakobi1997evolutionary}
Nick Jakobi.
\newblock Evolutionary robotics and the radical envelope-of-noise hypothesis.
\newblock {\em Adaptive behavior}, 6(2):325--368, 1997.

\bibitem[\protect\citeauthoryear{Jiang \bgroup \em et al.\egroup }{2021a}]{jiang2021replay}
Minqi Jiang, Michael Dennis, Jack Parker-Holder, Jakob Foerster, Edward Grefenstette, and Tim Rockt{\"a}schel.
\newblock Replay-guided adversarial environment design.
\newblock {\em Advances in Neural Information Processing Systems}, 34:1884--1897, 2021.

\bibitem[\protect\citeauthoryear{Jiang \bgroup \em et al.\egroup }{2021b}]{jiang2021prioritized}
Minqi Jiang, Edward Grefenstette, and Tim Rockt{\"a}schel.
\newblock Prioritized level replay.
\newblock In {\em International Conference on Machine Learning}, pages 4940--4950. PMLR, 2021.

\bibitem[\protect\citeauthoryear{Kiran \bgroup \em et al.\egroup }{2021}]{kiran2021deep}
B~Ravi Kiran, Ibrahim Sobh, Victor Talpaert, Patrick Mannion, Ahmad~A Al~Sallab, Senthil Yogamani, and Patrick P{\'e}rez.
\newblock Deep reinforcement learning for autonomous driving: A survey.
\newblock {\em IEEE Transactions on Intelligent Transportation Systems}, 23(6):4909--4926, 2021.

\bibitem[\protect\citeauthoryear{Liu \bgroup \em et al.\egroup }{2024}]{liu2024llm}
Jijia Liu, Chao Yu, Jiaxuan Gao, Yuqing Xie, Qingmin Liao, Yi~Wu, and Yu~Wang.
\newblock Llm-powered hierarchical language agent for real-time human-ai coordination.
\newblock In {\em Proceedings of the 23rd International Conference on Autonomous Agents and Multiagent Systems}, pages 1219--1228, 2024.

\bibitem[\protect\citeauthoryear{Lou \bgroup \em et al.\egroup }{2023}]{lou2023pecan}
Xingzhou Lou, Jiaxian Guo, Junge Zhang, Jun Wang, Kaiqi Huang, and Yali Du.
\newblock Pecan: Leveraging policy ensemble for context-aware zero-shot human-ai coordination.
\newblock In {\em Proceedings of the 2023 International Conference on Autonomous Agents and Multiagent Systems}, pages 679--688, 2023.

\bibitem[\protect\citeauthoryear{Lupu \bgroup \em et al.\egroup }{2021}]{lupu2021trajectory}
Andrei Lupu, Brandon Cui, Hengyuan Hu, and Jakob Foerster.
\newblock Trajectory diversity for zero-shot coordination.
\newblock In {\em International conference on machine learning}, pages 7204--7213. PMLR, 2021.

\bibitem[\protect\citeauthoryear{Oroojlooy and Hajinezhad}{2023}]{oroojlooy2023review}
Afshin Oroojlooy and Davood Hajinezhad.
\newblock A review of cooperative multi-agent deep reinforcement learning.
\newblock {\em Applied Intelligence}, 53(11):13677--13722, 2023.

\bibitem[\protect\citeauthoryear{Parker-Holder \bgroup \em et al.\egroup }{2022}]{parker2022evolving}
Jack Parker-Holder, Minqi Jiang, Michael Dennis, Mikayel Samvelyan, Jakob Foerster, Edward Grefenstette, and Tim Rockt{\"a}schel.
\newblock Evolving curricula with regret-based environment design.
\newblock In {\em International Conference on Machine Learning}, pages 17473--17498. PMLR, 2022.

\bibitem[\protect\citeauthoryear{Ruhdorfer \bgroup \em et al.\egroup }{2024}]{ruhdorfer2024overcooked}
Constantin Ruhdorfer, Matteo Bortoletto, Anna Penzkofer, and Andreas Bulling.
\newblock The overcooked generalisation challenge.
\newblock {\em arXiv preprint arXiv:2406.17949}, 2024.

\bibitem[\protect\citeauthoryear{Rutherford \bgroup \em et al.\egroup }{2025}]{rutherford2025no}
Alexander Rutherford, Michael Beukman, Timon Willi, Bruno Lacerda, Nick Hawes, and Jakob Foerster.
\newblock No regrets: Investigating and improving regret approximations for curriculum discovery.
\newblock {\em Advances in Neural Information Processing Systems}, 37:16071--16101, 2025.

\bibitem[\protect\citeauthoryear{Samvelyan \bgroup \em et al.\egroup }{2023}]{samvelyan2023maestro}
Mikayel Samvelyan, Akbir Khan, Michael~D Dennis, Minqi Jiang, Jack Parker-Holder, Jakob~Nicolaus Foerster, Roberta Raileanu, and Tim Rockt{\"a}schel.
\newblock {MAESTRO}: Open-ended environment design for multi-agent reinforcement learning.
\newblock In {\em The Eleventh International Conference on Learning Representations}, 2023.

\bibitem[\protect\citeauthoryear{Sarkar \bgroup \em et al.\egroup }{2023}]{sarkar2023diverse}
Bidipta Sarkar, Andy Shih, and Dorsa Sadigh.
\newblock Diverse conventions for human-ai collaboration.
\newblock {\em Advances in Neural Information Processing Systems}, 36:23115--23139, 2023.

\bibitem[\protect\citeauthoryear{Schulman \bgroup \em et al.\egroup }{2017}]{schulman2017proximal}
John Schulman, Filip Wolski, Prafulla Dhariwal, Alec Radford, and Oleg Klimov.
\newblock Proximal policy optimization algorithms.
\newblock {\em arXiv preprint arXiv:1707.06347}, 2017.

\bibitem[\protect\citeauthoryear{Silver \bgroup \em et al.\egroup }{2017}]{silver2017mastering}
David Silver, Thomas Hubert, Julian Schrittwieser, Ioannis Antonoglou, Matthew Lai, Arthur Guez, Marc Lanctot, Laurent Sifre, Dharshan Kumaran, Thore Graepel, et~al.
\newblock Mastering chess and shogi by self-play with a general reinforcement learning algorithm.
\newblock {\em arXiv preprint arXiv:1712.01815}, 2017.

\bibitem[\protect\citeauthoryear{Stone \bgroup \em et al.\egroup }{2010}]{stone2010ad}
Peter Stone, Gal Kaminka, Sarit Kraus, and Jeffrey Rosenschein.
\newblock Ad hoc autonomous agent teams: Collaboration without pre-coordination.
\newblock In {\em Proceedings of the AAAI Conference on Artificial Intelligence}, volume~24, pages 1504--1509, 2010.

\bibitem[\protect\citeauthoryear{Strouse \bgroup \em et al.\egroup }{2021}]{strouse2021collaborating}
DJ~Strouse, Kevin McKee, Matt Botvinick, Edward Hughes, and Richard Everett.
\newblock Collaborating with humans without human data.
\newblock {\em Advances in Neural Information Processing Systems}, 34:14502--14515, 2021.

\bibitem[\protect\citeauthoryear{Vinyals \bgroup \em et al.\egroup }{2019}]{vinyals2019alphastar}
Oriol Vinyals, Igor Babuschkin, Junyoung Chung, Michael Mathieu, Max Jaderberg, Wojciech~M Czarnecki, Andrew Dudzik, Aja Huang, Petko Georgiev, Richard Powell, et~al.
\newblock Alphastar: Mastering the real-time strategy game starcraft ii.
\newblock {\em DeepMind blog}, 2:20, 2019.

\bibitem[\protect\citeauthoryear{Wang \bgroup \em et al.\egroup }{2020}]{wang2020enhanced}
Rui Wang, Joel Lehman, Aditya Rawal, Jiale Zhi, Yulun Li, Jeffrey Clune, and Kenneth Stanley.
\newblock Enhanced poet: Open-ended reinforcement learning through unbounded invention of learning challenges and their solutions.
\newblock In {\em International Conference on Machine Learning}, pages 9940--9951. PMLR, 2020.

\bibitem[\protect\citeauthoryear{Yan \bgroup \em et al.\egroup }{2023}]{yan2023efficient}
Xue Yan, Jiaxian Guo, Xingzhou Lou, Jun Wang, Haifeng Zhang, and Yali Du.
\newblock An efficient end-to-end training approach for zero-shot human-ai coordination.
\newblock {\em Advances in Neural Information Processing Systems}, 36:2636--2658, 2023.

\bibitem[\protect\citeauthoryear{Yang \bgroup \em et al.\egroup }{2022}]{yang2022optimal}
Mesut Yang, Micah Carroll, and Anca Dragan.
\newblock Optimal behavior prior: Data-efficient human models for improved human-ai collaboration.
\newblock {\em arXiv preprint arXiv:2211.01602}, 2022.

\bibitem[\protect\citeauthoryear{Yuan \bgroup \em et al.\egroup }{2023}]{yuan2023learning}
Lei Yuan, Lihe Li, Ziqian Zhang, Feng Chen, Tianyi Zhang, Cong Guan, Yang Yu, and Zhi-Hua Zhou.
\newblock Learning to coordinate with anyone.
\newblock In {\em Proceedings of the Fifth International Conference on Distributed Artificial Intelligence}, pages 1--9, 2023.

\bibitem[\protect\citeauthoryear{Zhang \bgroup \em et al.\egroup }{2021}]{zhang2021multi}
Kaiqing Zhang, Zhuoran Yang, and Tamer Ba{\c{s}}ar.
\newblock Multi-agent reinforcement learning: A selective overview of theories and algorithms.
\newblock {\em Handbook of reinforcement learning and control}, pages 321--384, 2021.

\bibitem[\protect\citeauthoryear{Zhao \bgroup \em et al.\egroup }{2023}]{zhao2023maximum}
Rui Zhao, Jinming Song, Yufeng Yuan, Haifeng Hu, Yang Gao, Yi~Wu, Zhongqian Sun, and Wei Yang.
\newblock Maximum entropy population-based training for zero-shot human-ai coordination.
\newblock In {\em Proceedings of the AAAI Conference on Artificial Intelligence}, volume~37, pages 6145--6153, 2023.

\end{thebibliography}

\clearpage
\appendix
\section{Implementation Details and Hyperparameters.}\label{sec_params}

Our implementation is based on PLR \cite{jiang2021prioritized} and Overcooked-AI \cite{carroll2019utility}. In all experiments, we train the ego-agent using the PPO method \cite{schulman2017proximal}. The details of the network architecture and hyperparameter choices are provided in Tables \ref{tab:network} and Table \ref{tab:ppo_agent_parameter}, respectively. Additionally, Table \ref{tab:each_method} presents the final hyperparameter choices for all methods.
 
\begin{table}[ht]
\caption{Network architecture details}
\label{tab:network}
\centering
\begin{tabular}{c l c c}
    \toprule
    \textbf{Layers} & \textbf{Channels} & \textbf{Kernel Size} \\
    \midrule
    Conv2D + LeakyReLU & 25 & 5$\times$5 \\
    Conv2D + LeakyReLU & 25 & 3$\times$3 \\
    Conv2D + LeakyReLU & 25 & 3$\times$3 \\
    Flatten & / & / \\
    (FC + LeakyReLU)$\times$3 & 64 & / \\
    FC & 6 & / \\
    \bottomrule
\end{tabular}
\label{tab:network_architecture}
\end{table}

\begin{table}[ht]
\centering
\caption{PPO agent hyperparameters}
\label{tab:ppo_agent_parameter}
\begin{tabular}{m{4cm}m{2cm}}
\toprule
\textbf{Parameter} & \textbf{Value} \\
\midrule
\textbf{PPO} & \\
$\lambda_{\text{GAE}}$ & 0.98 \\
$\gamma$ & 0.99 \\
Number of epochs & 8 \\
Rollout Length & 400 \\
PPO clip range & 0.05 \\
RMSprop optimizer ($\epsilon$) & 0.00001 \\
Learning rate & 0.001 \\
Value loss coefficient & 0.1 \\
Entropy Coefficient & 0.1 \\
Number of mini-batches & 20 \\
Minibatch size & 5000 \\
\bottomrule
\end{tabular}
\end{table}

\noindent\textbf{Overcooked-AI.} To evaluate the improvement in zero-shot human AI coordination performance in unseen environments, we experiment in the Overcooked AI environment, where players can cook and serve food. The environment has interactive blocks such as onions, plates, pots, and outlets, wall blocks that agents cannot move through, and space blocks that agents can move through. To get a reward from the environment, each player picks up 3 onions (3 reward) and puts them in the pot (3 reward). After 20 seconds, the onion soup is ready. Take a plate to serve the onion soup (5 reward). They then submit the dish to the outlets. Finally, all players get a common reward of 20. This reward shaping is based on a previous Overcooked-AI benchmark paper \cite{carroll2019utility}.

\begin{table}[ht]
\centering
\caption{Training hyperparameters for each method}
\label{tab:each_method}
\begin{tabular}{m{4cm}m{3cm}}
\toprule
\textbf{Parameter} & \textbf{Value} \\
\midrule
\textbf{Robust PLR} & \\
Replay rate, $p$ & 0.5 \\
Buffer size, $K$ & 4,000 \\
Scoring function & Positive value loss \\
Prioritization & rank \\
Temperature, $\beta$ & 0.3 \\
Staleness coefficient, $\rho$ & 0.3 \\
\midrule
\textbf{MAESTRO} & \\
$\lambda$ coef & 0.2 \\
Buffer size of co-player, $K$ & 1,000 \\
Scoring function & Positive value loss  \\
Prioritization & rank \\
Population size, $i$ & 8 \\
Episodes $N$ & 375 \\  
\midrule
\textbf{Our} & \\
$\lambda$ coef & 0.2 \\
Buffer size of co-player, $K$ & 1,000 \\
Scoring function & return \\
Prioritization & reverse-rank \\
Population size, $i$ & 8 \\
% Dissimilarity coefficient, $\nu$ & 0.1\\
Episodes $N$ & 375 \\  
\bottomrule
\end{tabular}
\end{table}

\section{Layout Generator.}\label{sec_layout_generator}

\begin{algorithm}[ht]
\caption{Layout generation process}
\textbf{Input:} Maximum size of layout buffer $M$, Number of interactive blocks $N$, Number of empty space $E$ \\
\textbf{Initialise:} Layout buffer $L$ \\
\vspace{-\baselineskip}
\begin{algorithmic}[1] % Enables numbering for each row
\WHILE{Size of layout buffer $<$ $M$}
    \STATE Make random $7\times5$ array as a Layout
    \STATE Randomly place $N$ blocks
    \IF {Duplicate(Layout, $L$)} 
        \STATE{Continue}
    \ENDIF
    \STATE Place two players in a random position
    \STATE Remove non-reachable blocks
    
    \IF {Solvability(Layout)}
        \STATE Append Layout to $L$
    \ENDIF
\ENDWHILE
\end{algorithmic}
\label{layout_algo}
\end{algorithm}

\begin{figure*}[ht]
    \centering
    \subfloat{\includegraphics[width=0.32\linewidth]{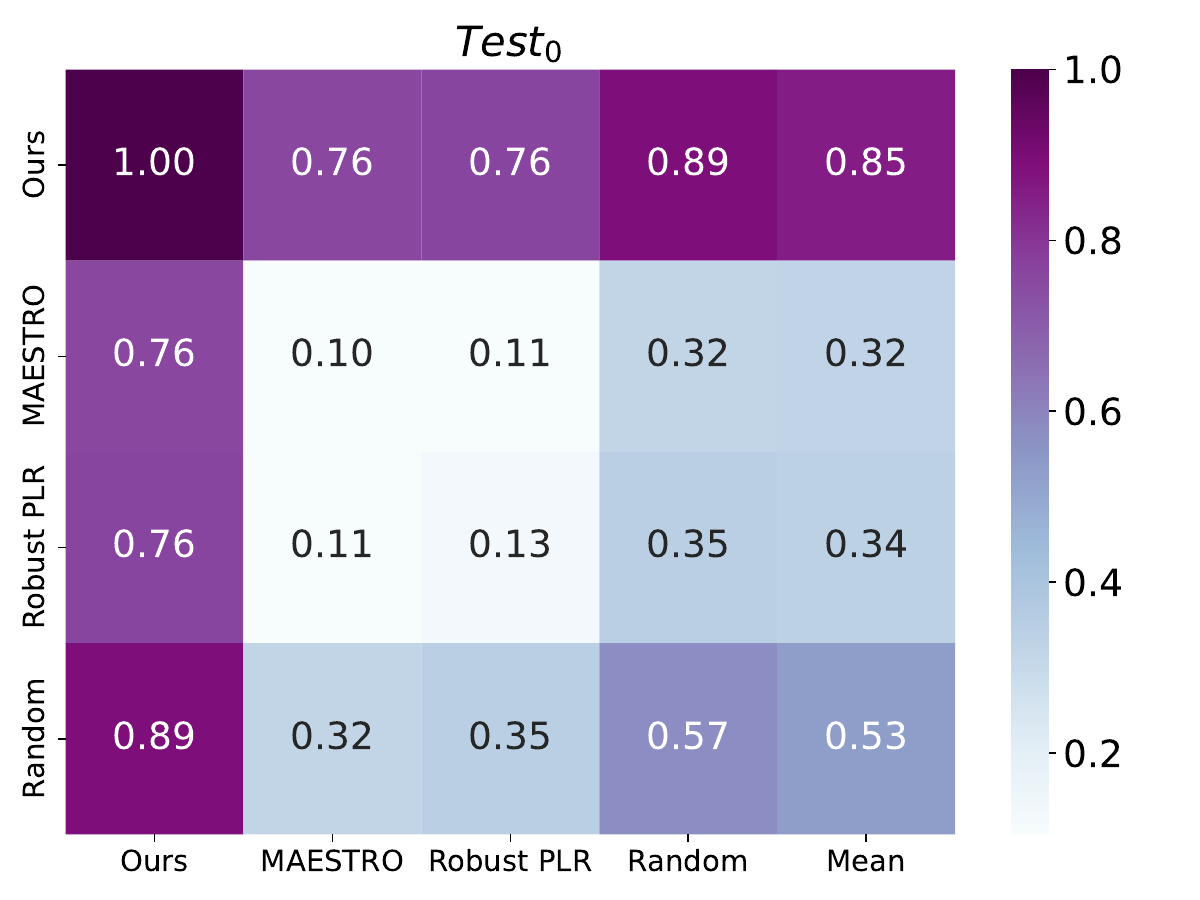}}
    \hspace{0.002\linewidth}
    \subfloat{\includegraphics[width=0.32\linewidth]{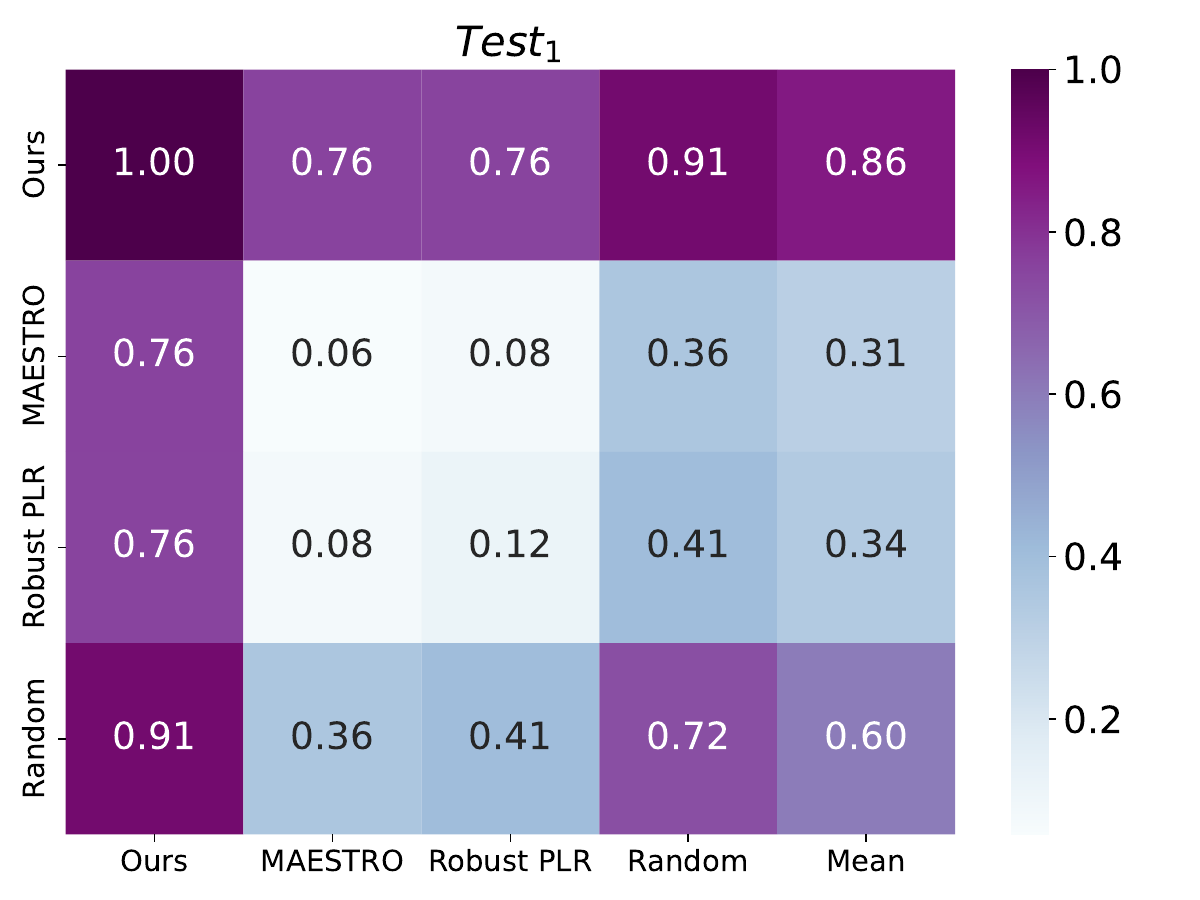}}
    \hspace{0.002\linewidth}
    \subfloat{\includegraphics[width=0.32\linewidth]{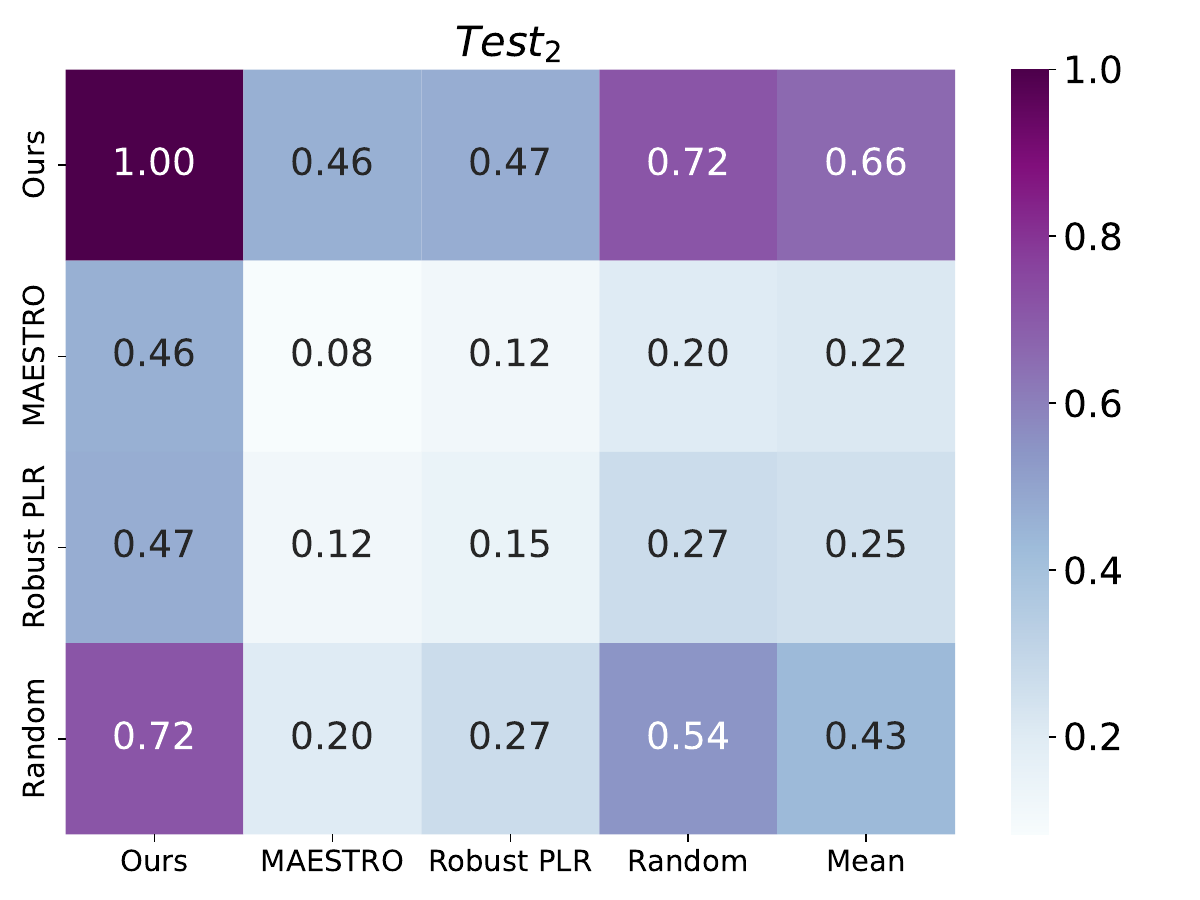}}
    \hspace{0.002\linewidth}
    \vspace*{1em}
    \subfloat{\includegraphics[width=0.32\linewidth]{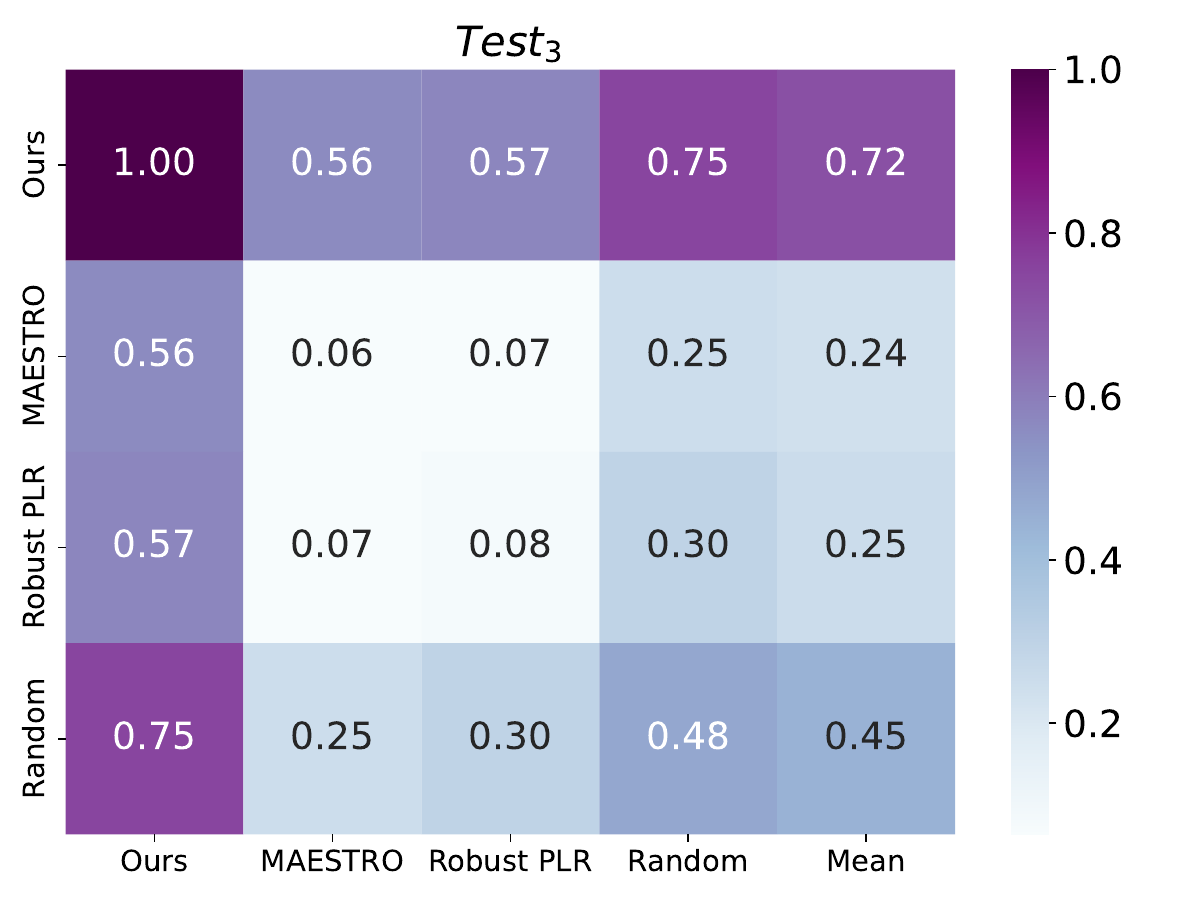}}
    \hspace{0.002\linewidth}
    \subfloat{\includegraphics[width=0.32\linewidth]{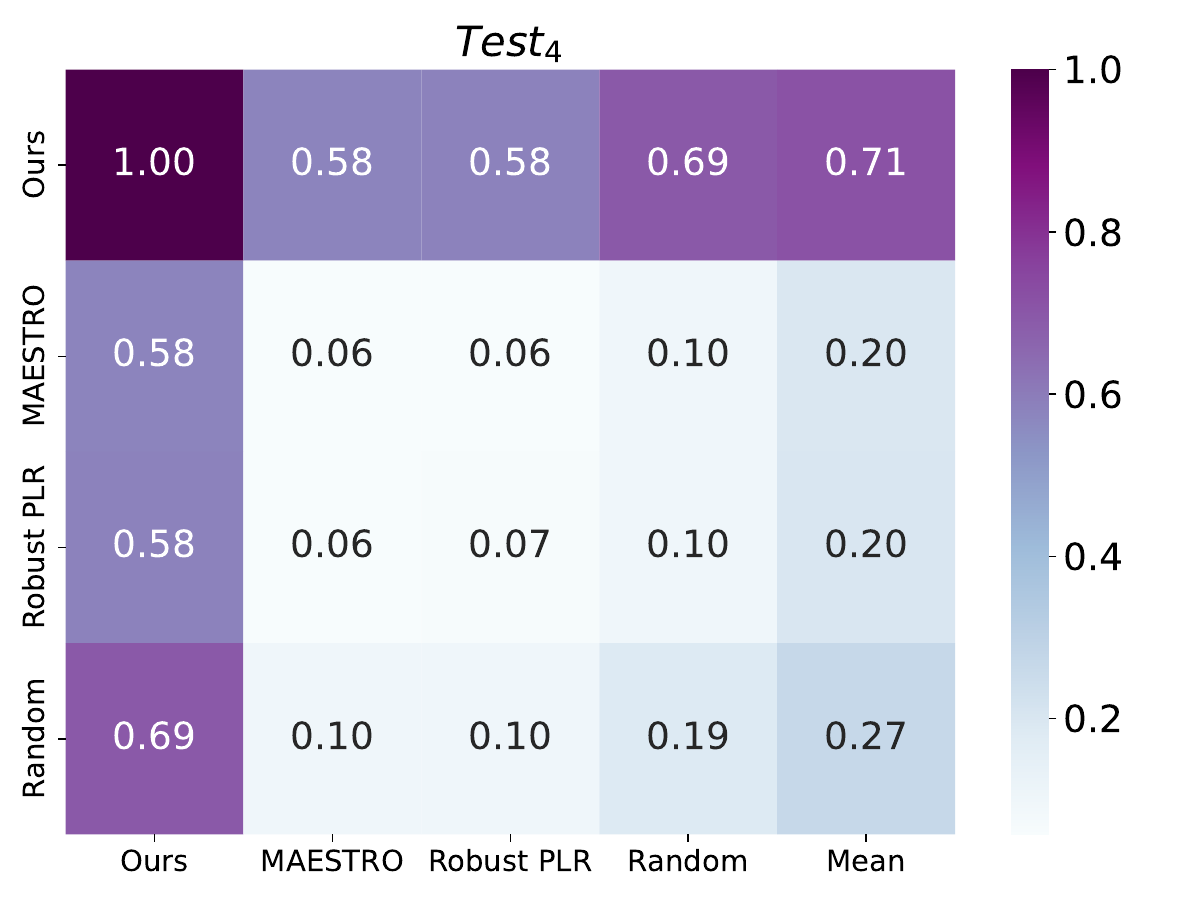}}
    \hspace{0.002\linewidth}
    \subfloat{\includegraphics[width=0.32\linewidth]{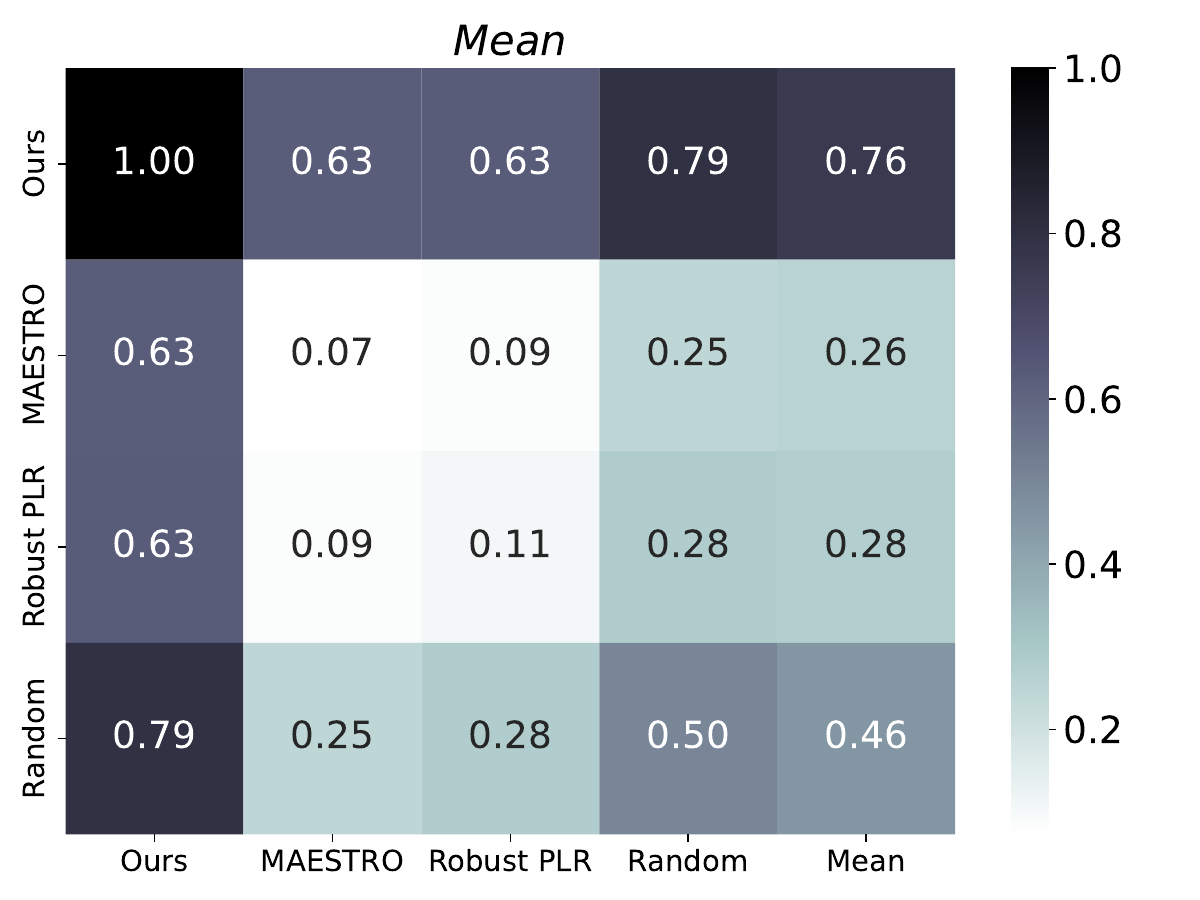}}
    \hspace{0.002\linewidth}
    \caption{\textbf{AI-AI Cross-play result} This result is the normalized value of the reward obtained by cross-playing the trained model on five evaluation layouts and their average.}
    \vspace{-5mm}
    \label{cross-play}
\end{figure*}

The algorithm \ref{layout_algo} shows the overall layout generation process. A layout generator is a heuristic-driven tool whose input consists of three parameters: The maximum size of the layout buffer $M$, the number of interactive blocks $N$, and the number of empty spaces $E$. The generator ensures the presence of a minimum of $E$ spaces and $N$ interactive blocks within each map. The locations of the blocks in the layout are randomly selected. In this experiment, we set the parameters as $M=6,000$, $N = 6 \sim 9$, and $E=14$. During the layout generation process, the generator checks the Hamming distance between the embedded arrays to eliminate any redundancy. Additionally, we verify the solvability of each layout using the A* path-finding algorithm \cite{hart1968formal}, which determines whether the agent can reach each interactive block.

\section{Cross-Play Results.}\label{sec_crossplay}

Figure \ref{cross-play} represents the detailed results of experiments with AI-AI cross-play on the five evaluation layouts. This is a zero-shot coordination experiment in which different models are coordinated without direct encounters with others during the training phase. These scores are normalized relative to the highest reward obtained from that layout, with the maximum value set at 1 and the minimum value set to 0. Our method performs better than the baselines in all five evaluation layouts. The final mean graph shows the average value of the normalized scores for the five layouts. Our method also outperforms baselines in self-play where each method plays itself, and in cross-play where each method plays against different methods.

Robust PLR and MAESTRO use a regret-based utility function, which is not suitable for cooperative environments, resulting in worse performance than the random method. In contrast, our method improves the lower bound of the performance by using a return-based utility function. This approach curates challenging environment/co-player pairs for the ego-agent. Our trained ego-agent generally performs well in easy-to-hard environments and is robust to unseen environments and co-players.

\section{ADDITIONAL RESULTS.}\label{sec_buffer}

\begin{table}[ht]
\caption{Difficulty classification based on reward conditions}
\centering
\begin{tabular}{ll}
\hline
\textbf{Difficulty} & \textbf{Reward Condition}  \\ \hline
Very Easy           & \( reward > \mu + 1.5\sigma \) \\
Easy                & \( \mu + 0.5\sigma < reward \leq \mu + 1.5\sigma \) \\
Medium              & \( \mu - 0.5\sigma < reward \leq \mu + 0.5\sigma \) \\
Hard                & \( \mu - 1.5\sigma < reward \leq \mu - 0.5\sigma \) \\
Very Hard           & \( reward \leq \mu - 1.5\sigma \) \\ \hline
\end{tabular}
\label{tab:difficulty_classification}
\vspace{-3mm}
\end{table}

\begin{figure*}[ht]
    \centering
    \begin{minipage}{\textwidth}
        \centering
        \resizebox{0.95\textwidth}{!}{
        \subfloat[Difficulty categorized by reward per layouts]{\includegraphics[width=0.45\linewidth]{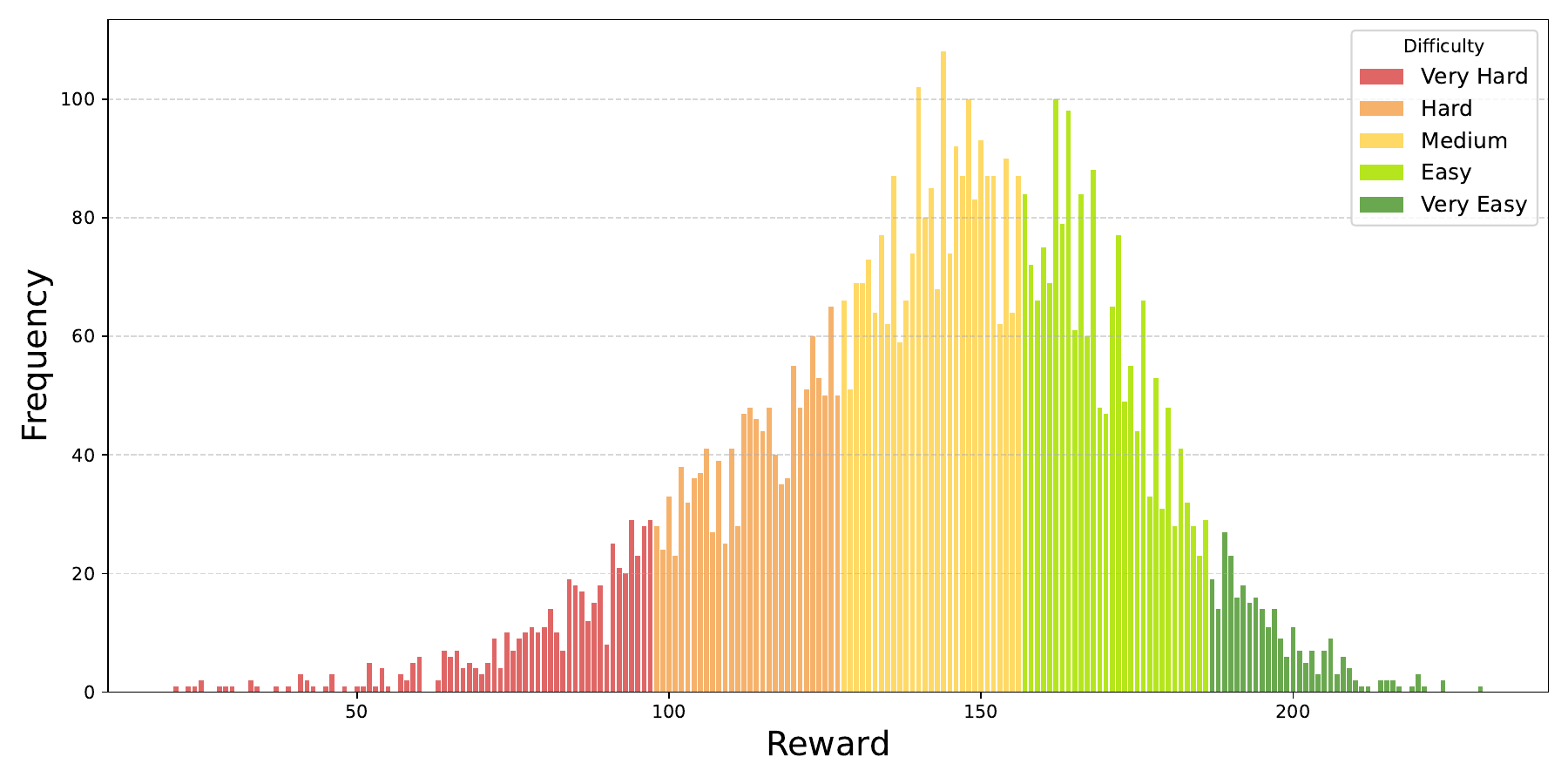}}
        \subfloat[Layout difficulty composition in the co-player's buffer]{\includegraphics[width=0.275\linewidth]{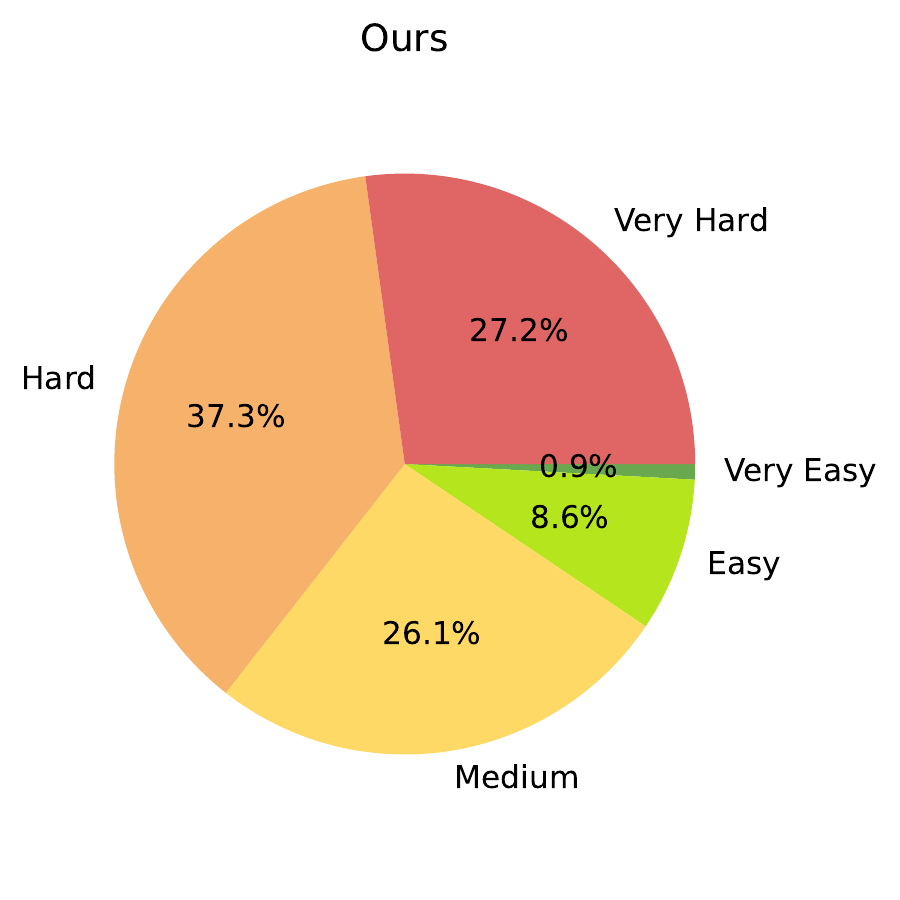}\includegraphics[width=0.275\linewidth]{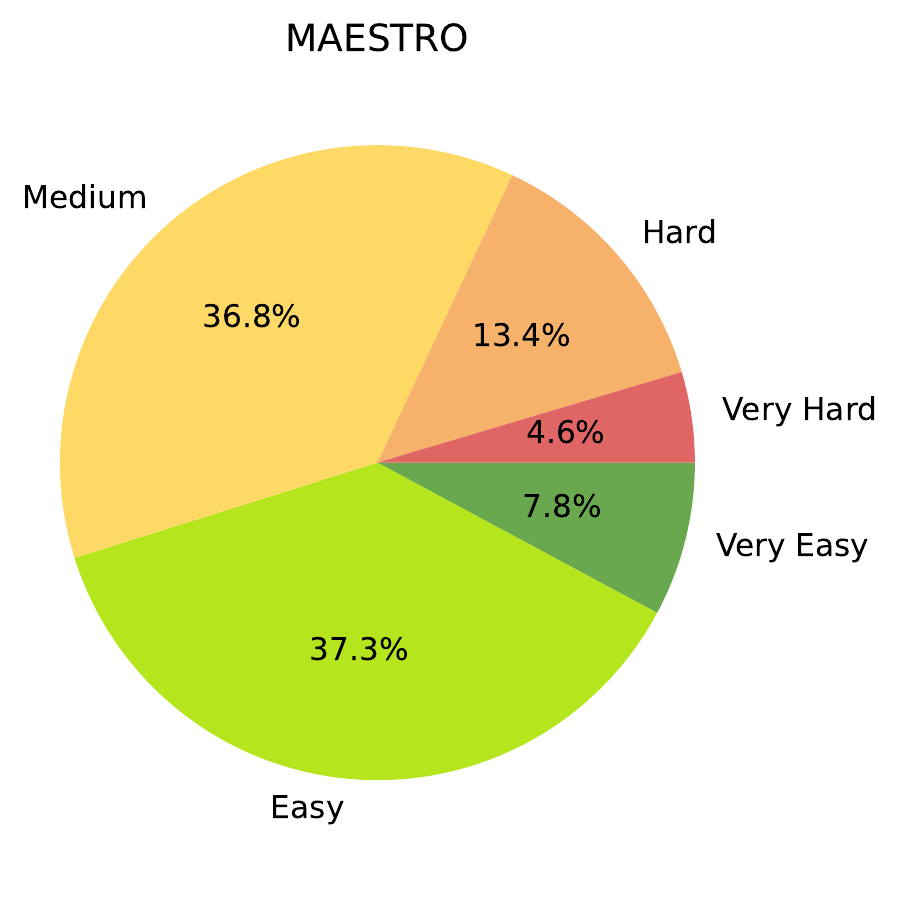}}
        }
        \subfloat[Changing the layout difficulty composition of buffers]{\includegraphics[width=0.48\linewidth]{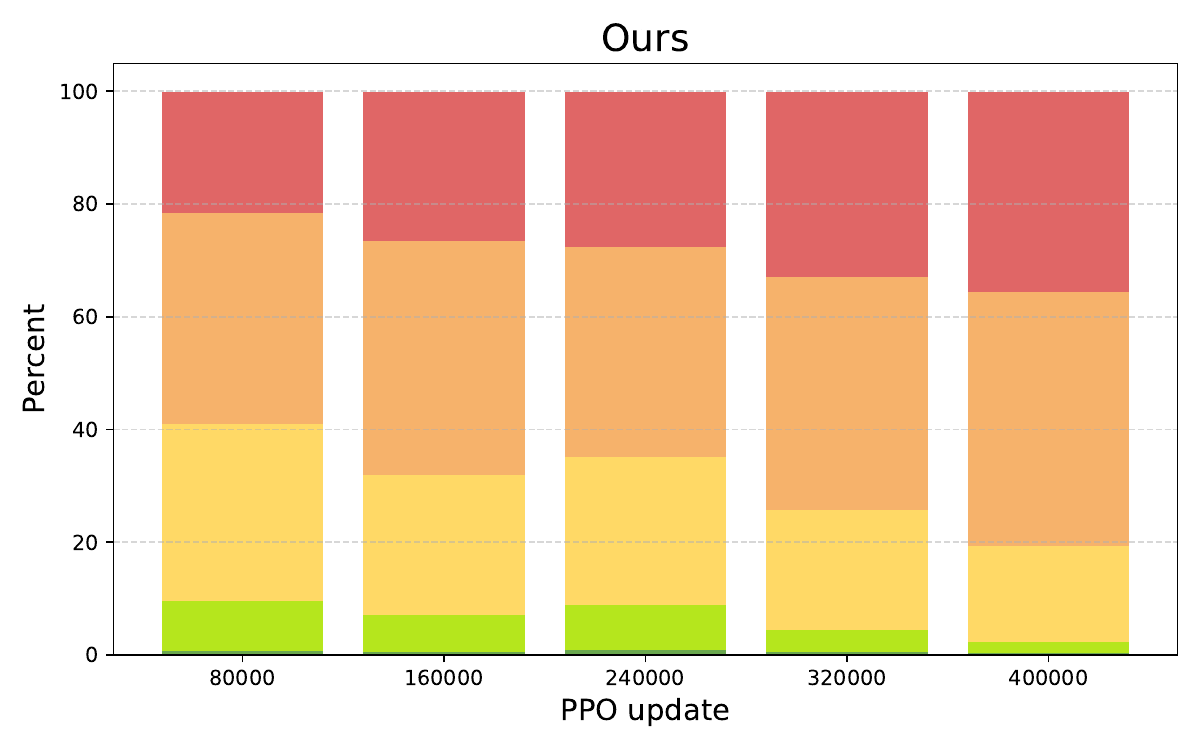}\includegraphics[width=0.48\linewidth]{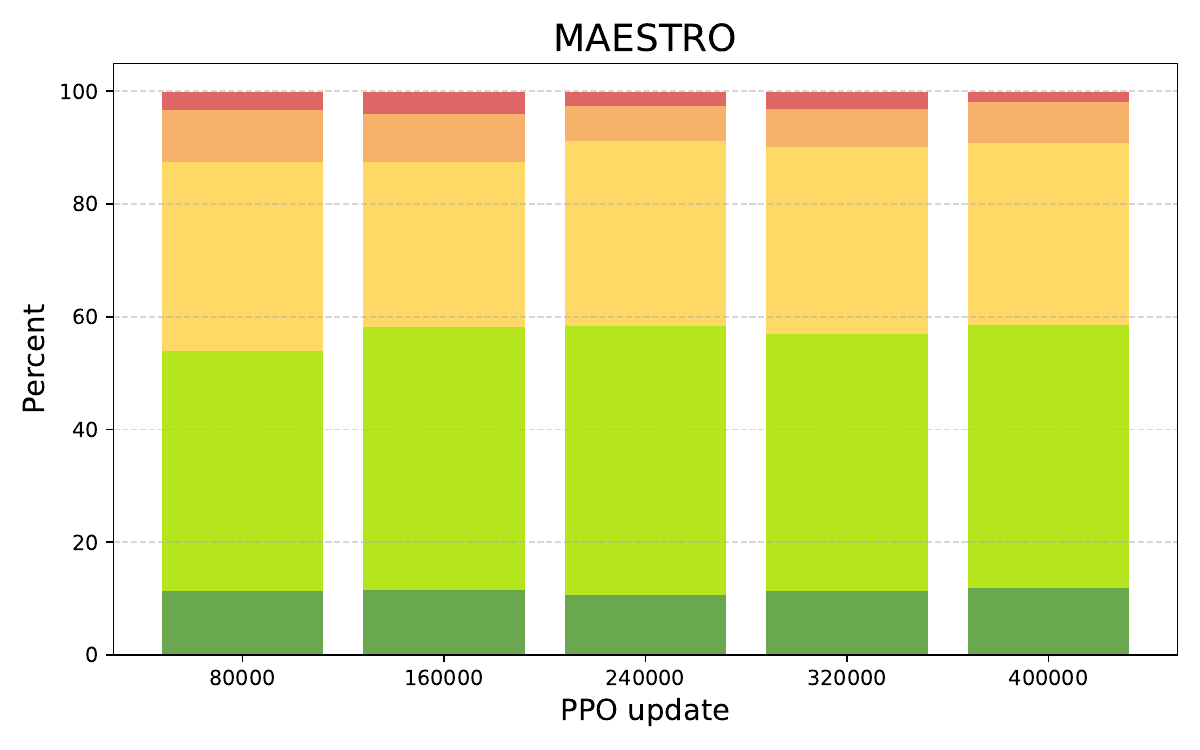}}
        \caption{(a) Distribution of rewards obtained by the trained model on training layouts, playing with a human proxy. Difficulty is categorized based on these reward distributions. (b) Difficulty distribution of layouts sampled during the training process for our method and MAESTRO. (c) The difficulty configuration of the buffer changes as training progresses.}
        \label{diff-dist}
    \end{minipage}

    \vspace{10mm}

    \begin{minipage}{\textwidth}
        \centering
         \resizebox{0.95\textwidth}{!}{
        \includegraphics[width=\textwidth]{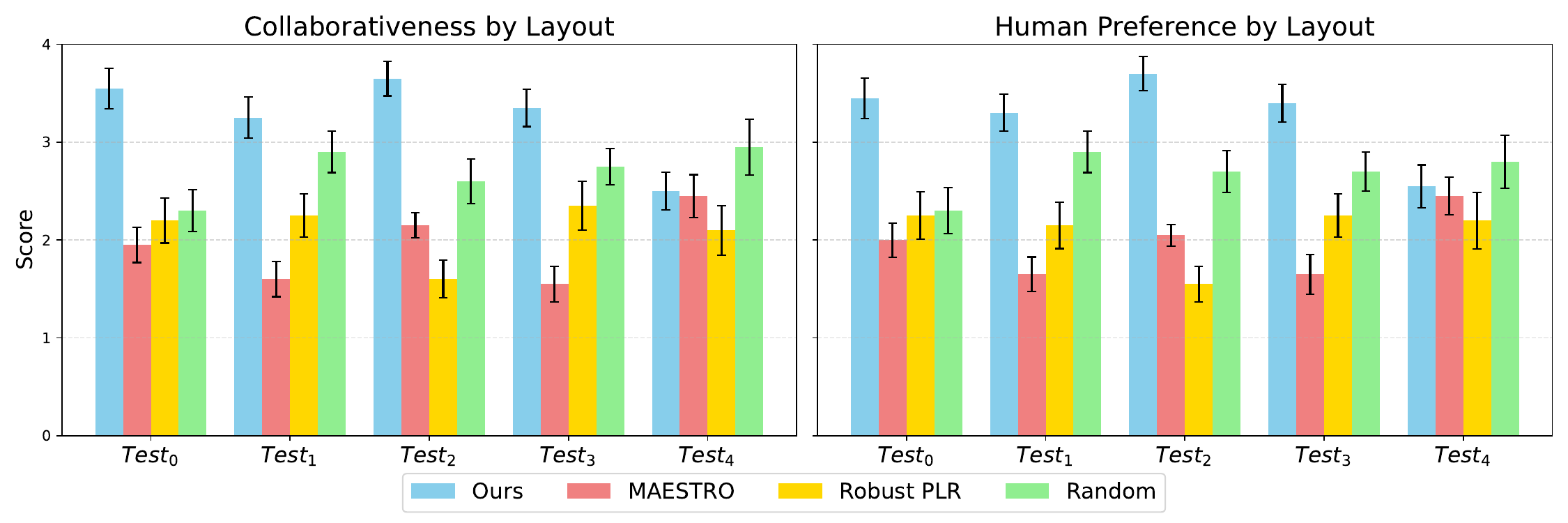}
        }
        \caption{\textbf{Human survey results by test layouts} Results from a survey of 20 human players on five evaluation layouts. In the layouts of \textit{$test_0, test_1, test_2$} and \textit{$test_3$}, human players tended to prefer our model, but in the hardest layout of \textit{$test_4$}, the human player had different preferences, and each model scored similarly.}
        \label{fig:human-ai_layout}
    \end{minipage}
\end{figure*}

\noindent\textbf{Define training layouts difficulty.}
To categorize the difficulty of the layouts, we obtained the average reward (50 runs) evaluated from the trained agent and the human proxy agent. Figure \ref{diff-dist}(a) shows the distribution of average rewards obtained from the 6,000 layouts used for training. We applied the criteria outlined in Table \ref{tab:difficulty_classification}, categorizing the difficulty levels into five groups, from 'Very Easy' to 'Very Hard', based on the mean and standard deviation of the reward distribution. According to this classification scheme, the layouts classified as 'Medium' are the most numerous, while those classified as 'Very Hard' or 'Very Easy' are comparatively rare.

\noindent\textbf{Co-player's buffer Composition.}
Figure \ref{diff-dist}(b) shows the cumulative composition of layout difficulties in the environment buffer of the co-player, which is first added to the co-player population during the training process. Both our method and MAESTRO prioritize layout(environment) sampling and store layouts in the co-player’s buffer of size 1,000 according to their own criteria. In our method, the percentage of 'Very Hard' layouts in the co-player’s buffer increases as the training progresses from start to finish. This indicates that our method samples the more difficult layouts as the agent's performance improves, similar to curriculum learning. However, MAESTRO was likely to sample easier environments as training progressed, so the percentage of 'Easy' layouts in the co-player buffer increased. Figure \ref{diff-dist}(c) represents details of the changing layout composition in the co-player's buffer. These results support the idea that the return-based utility function is more appropriate than the regret-based utility function to assess learning potential in a coordination setting.

\section{ADDITIONAL Paired t-test.}\label{sec_t_test}

\begin{table}[h]
\caption{Paired t-test results between Ours and baseline methods on each evaluation layout with a human proxy.}
\label{tab:t_test_human_proxy}
\centering
\begin{tabular}{lccc}
\toprule
\textbf{Layout} & \textbf{Comparison} & \textbf{t-statistic} & \textbf{p-value} \\
\midrule
\multirow{3}{*}{$test_{0}$}
& MAESTRO & 4.8088 & 0.0406$^{*}$ \\
& Robust PLR & 4.4569 & 0.0468$^{*}$ \\
& Random     & 4.9371 & 0.0387$^{*}$ \\
\midrule
\multirow{3}{*}{$test_{1}$}
& MAESTRO & 29.0707 & 0.0012$^{**}$ \\
& Robust PLR & 9.4667 & 0.0110$^{*}$ \\
& Random     & 10.1263 & 0.0096$^{**}$ \\
\midrule
\multirow{3}{*}{$test_{2}$}
& MAESTRO & 7.6474 & 0.0167$^{*}$ \\
& Robust PLR & 207.0434 & $<$0.0001$^{***}$ \\
& Random     & 25.8937 & 0.0015$^{**}$ \\
\midrule
\multirow{3}{*}{$test_{3}$}
& MAESTRO & 17.2566 & 0.0033$^{**}$ \\
& Robust PLR & 10.8919 & 0.0083$^{**}$ \\
& Random     & 4.1389 & 0.0537 \\
\midrule
\multirow{3}{*}{$test_{4}$}
& MAESTRO & 3.5168 & 0.0722 \\
& Robust PLR & 4.0035 & 0.0571 \\
& Random     & 1.7264 & 0.2264 \\
\bottomrule
\end{tabular}
\end{table}

\vspace{-5pt}

\begin{table}[h]
\caption{Paired t-test results between Ours and baseline methods on each layout evaluated with a human partner.}
\label{tab:t_test_human}
\centering
\begin{tabular}{lccc}
\toprule
\textbf{Layout} & \textbf{Method} & \textbf{t-statistic} & \textbf{p-value} \\
\midrule
\multirow{3}{*}{$test_{0}$}
& MAESTRO & 3.2151 & 0.0046$^{**}$ \\
& Robust\_PLR & 3.6306 & 0.0018$^{**}$ \\
& Random & 2.9939 & 0.0075$^{**}$ \\
\midrule
\multirow{3}{*}{$test_{1}$}
& MAESTRO & 6.1094 & <0.0001$^{***}$ \\
& Robust\_PLR & 3.3245 & 0.0036$^{**}$ \\
& Random & 0.9126 & 0.3729 \\
\midrule
\multirow{3}{*}{$test_{2}$}
& MAESTRO & 4.8766 & 0.0001$^{***}$ \\
& Robust\_PLR & 5.6865 & <0.0001$^{***}$ \\
& Random & 3.9149 & 0.0009$^{***}$ \\
\midrule
\multirow{3}{*}{$test_{3}$}
& MAESTRO & 4.3184 & 0.0004$^{***}$  \\
& Robust\_PLR & 3.3836 & 0.0031$^{**}$ \\
& Random & 4.2080 & 0.0005$^{***}$ \\
\midrule
\multirow{3}{*}{$test_{4}$}
& MAESTRO & -0.4229 & 0.6771 \\
& Robust\_PLR & 0.5977 & 0.5571 \\
& Random & -1.6296 & 0.1197 \\
\bottomrule
\end{tabular}
\end{table}

\end{document}